\documentclass[sigconf]{acmart}
\AtBeginDocument{%
  \providecommand\BibTeX{{%
    \normalfont B\kern-0.5em{\scshape i\kern-0.25em b}\kern-0.8em\TeX}}}

\usepackage{multicol}
\usepackage{multirow}
\usepackage{array}
\usepackage{cleveref}
\usepackage{subcaption}
\usepackage{xcolor}
\usepackage{stackengine}
\usepackage{multicol}
\usepackage{multirow}
\usepackage{enumitem}
\usepackage{soul}

\usepackage{xcolor}
\usepackage{framed}

\newenvironment{myjsonblock}[1]{%
  \begin{center}%
    \noindent\colorbox{black}{%
      \parbox{0.95\textwidth}{%
        \strut\textcolor{white}{\bfseries #1}%
      }%
    }\\[6pt]%
    \begin{minipage}{0.95\textwidth}%
      \begin{framed}%
        \vspace*{6pt}%
}{%
        \vspace*{6pt}%
      \end{framed}%
    \end{minipage}%
  \end{center}%
}

\usepackage{multirow}
\usepackage{subcaption}
\definecolor{ForestGreen}{RGB}{34,139,34}
\newcommand{\corpus}{LAMP Corpus}

\usepackage{pifont}% http://ctan.org/pkg/pifont

\begin{document}

\title{Can AI writing be salvaged? Mitigating Idiosyncrasies and Improving Human-AI Alignment in the Writing Process through Edits}

\author{Tuhin Chakrabarty}
\email{tchakrabarty@salesforce.com}
\affiliation{%
  \institution{Salesforce AI Research}
  \country{USA}
}

\author{Philippe Laban}
\affiliation{%
  \institution{Salesforce AI Research}
  \country{USA}
  \email{}
}

\author{Chien-Sheng Wu}
\affiliation{%
  \institution{Salesforce AI Research}
  \country{USA}
}

\renewcommand{\shortauthors}{Chakrabarty, et al.}

\begin{abstract}
LLM-based applications are helping people write, and LLM-generated text is making its way into social media, journalism, and our classrooms. However, the differences between LLM-generated and human-written text remain unclear. To explore this, we hired professional writers to edit paragraphs in several creative domains. We first found these writers agree on undesirable idiosyncrasies in LLM-generated text, formalizing it into a seven-category taxonomy (e.g. clichés, unnecessary exposition). Second, we curated the LAMP corpus: 1,057 LLM-generated paragraphs edited by professional writers according to our taxonomy. Analysis of LAMP reveals that none of the LLMs used in our study (GPT4o, Claude-3.5-Sonnet, Llama-3.1-70b) outperform each other in terms of writing quality, revealing common limitations across model families. Third, building on existing work in automatic editing we evaluated methods to improve LLM-generated text. A large-scale preference annotation confirms that although experts largely prefer text edited by other experts, automatic editing methods show promise in improving alignment between LLM-generated and human-written text.
\end{abstract}



% Author Keywords


\maketitle

\section{Introduction}

Artificial Intelligence (AI) has the potential to revolutionize how we write, communicate, and express ideas~\citep{10.1145/3613904.3642625}. Recent studies have demonstrated the potential of large language models (LLMs) in assisting with various writing tasks, including argumentative \citep{lee2022coauthor,10.1145/3613904.3642134}, scientific \citep{gero2022sparks}, and creative writing \citep{10.1145/3635636.3656201,ippolito2022creative,yuan2022wordcraft,mirowski2023cowriting,mirowski2024robot}. Aligning LLMs with human preferences \citep{ouyang2022training} has enabled their transformation into user-friendly tools for non-technical users, such as Google's WorkSpace Labs, Grammarly, and Sudowrite. However, to truly benefit society, AI writing assistants must enhance human creativity and expression rather than homogenize content or diminish linguistic diversity \citep{gabriel2024ethics,10.1145/3544548.3581196}

\begin{figure*}
\centering
\includegraphics[width=\textwidth]{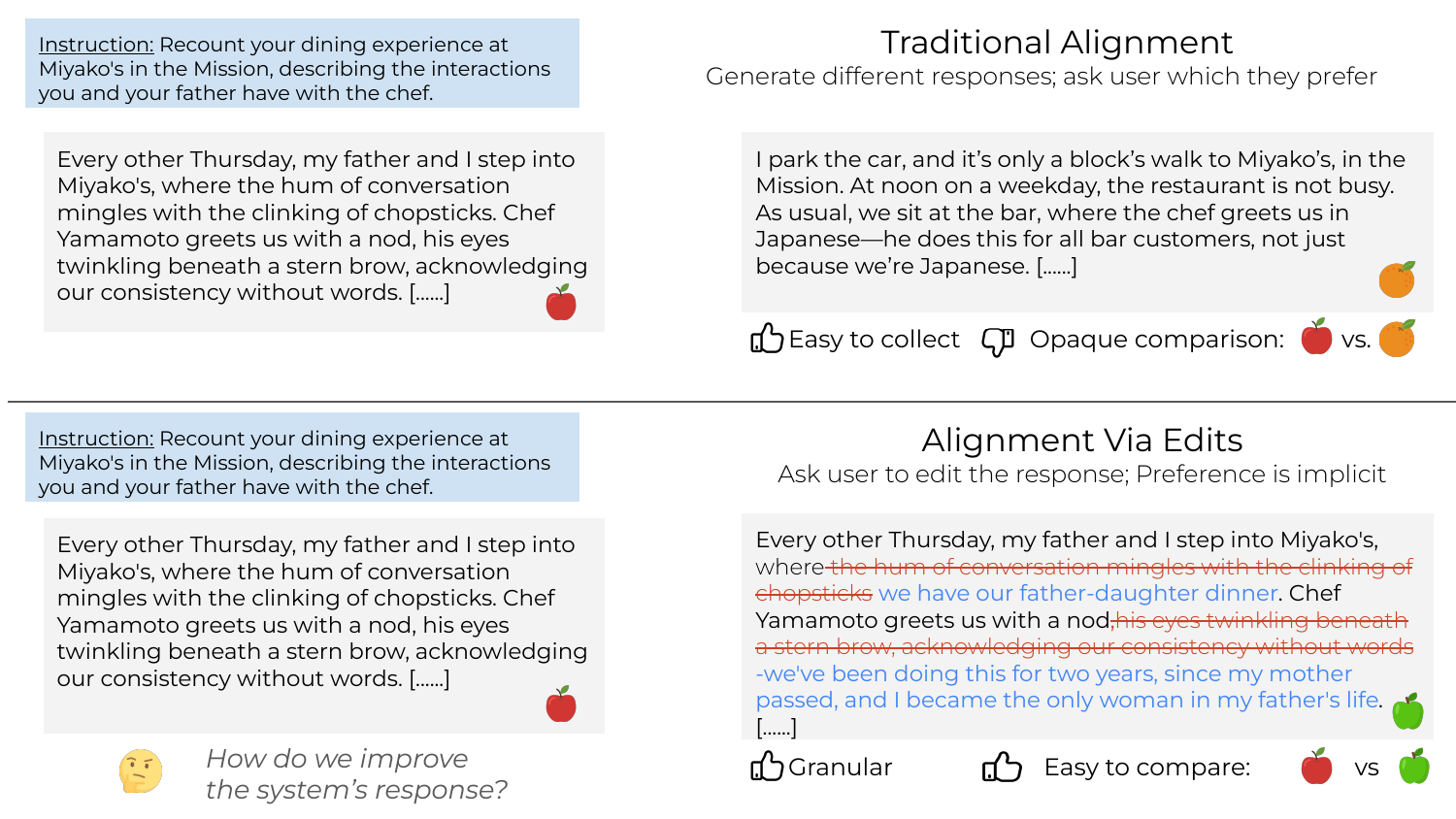}
\caption{\label{intro_fig} To align models to human preferences, human annotators are typically shown two responses and asked to choose the one they prefer. (i) The top portion of the Figure shows Traditional Alignment: it is often hard to compare two responses that differ widely. (ii) The bottom portion of the Figure shows Alignment via Edits where the original response is edited, allowing for a more granular comparison, with the edited version of the text naturally preferred over the original response.}
\end{figure*}

While LLM-based writing assistants have the potential to improve writing quality and increase author productivity, they also introduce an algorithmic monoculture \citep{kleinberg2021algorithmic}. \citet{padmakumar2023does} and \citet{10.1145/3635636.3656204} discuss how writing with LLMs unintentionally reduces content diversity, leading to homogenization. This homogenization occurs not just at the semantic level but also at syntactic (structural), lexical \cite{kobak2024delving}, and stylistic levels \citep{shaib2024detection}. For instance, prior work from \citet{ippolito2022creative,10.1145/3635636.3656201} has shown how LLM-generated text is often hackneyed and rife with clichés while failing to demonstrate rhetorical complexity and often revealing the subtext \textemdash a phenomenon known as ``telling instead of showing''~\citep{10.1145/3613904.3642731}. Additionally, LLM-generated texts are typically full of redundant exposition, overwrought metaphors, and florid descriptions due to verbosity bias during preference labeling~\citep{saito2023verbosity}. 

Current AI-assisted writing tools are powered by pre-trained language models that are refined through reinforcement learning from human feedback (RLHF) \cite{ziegler2019fine}. RLHF transforms human preferences into training data to guide language models toward desired outcomes. The most common type of feedback used with RLHF is binary preferences between pairs of examples sampled from one or more language models \cite{christiano2017deep} (See Fig \ref{intro_fig} Traditional Alignment). However, this approach has a drawback. The paired outputs may differ in numerous ways and could be equally flawed in containing idiosyncrasies. Asking an annotator to choose between two undesirable outputs does not improve alignment\footnote{In the current design of RLHF annotators are not allowed to not pick either}. \cite{casper2023open,ghosh2024compare}. We argue that alignment training needs to be aware of how desirable any individual response is, regardless of its preference relationship. Editing undesirable portions of a response can be seen as an effective mechanism for enhancing alignment (See Fig \ref{intro_fig} Alignment via Edits). An LLM-generated response that has been edited typically contains fewer undesirable traits and can be paired with the original LLM-generated response for preference ranking ($edited > original$). The challenge, however, lies in consistently identifying and implementing edits that enhance quality while aligning with human preferences for effective writing. Prompting techniques that encourage the model to self-edit \cite{madaan2024self} have shown promise; however, they do not work for long-form or paragraph-level writing \cite{pan2024spontaneous}. The primary reason for this is that LLMs do not inherently know what aspects of the writing need improvement, how extensively they should edit, or how to make changes that align with any given writer's expertise level and style.

To address these problems, we first create a comprehensive taxonomy of edit categories based on expert writing practices. We then recruit 18 writers to edit LLM-generated text using categories from our taxonomy. We define edits as changes that alter, replace, or refine specific phrases, clauses, or sentences within a larger text. We restrict our focus to generating text in literary fiction and creative non-fiction, as these genres challenge LLMs with their creativity, emotional nuance, and sophisticated language use.
We focus on paragraph-level edits, as they balance granularity and scope, reducing costs and annotator fatigue. Paragraphs capture style and context better than sentences, enabling more cohesive improvements. Given LLMs' limitations in long-term discourse coherence, paragraph-level enhancements facilitate human-AI collaboration. This approach allows humans to guide the overall structure and flow, while AI handles the lower-level details \cite{10.1145/3613904.3642134}. Finally, while expert writer's edits are valuable for identifying idiosyncrasies, this approach may not scale for large text volumes. To address this, we design few-shot prompts that use writer's edits to identify problematic spans in LLM-generated paragraphs and suggest improvements. This method aims to enhance overall paragraph quality at scale. Our work adds to the existing rich literature \cite{liu2022second,jiang2022arxivedits,gao2024aligning,yao2023improving} on using completion edits to improve alignment in different tasks. To summarize our contributions:

\begin{itemize}
    \item We propose a comprehensive edit taxonomy informed by expert writing practices that serve as a useful framework to identify and mitigate the idiosyncrasies in LLM-generated text.
    
    \item We release the LAMP (Language model Authored, Manually Polished) corpus containing 1057 <instructions,\\response> pairs grounded in real-world writing tasks such as Literary Fiction or Creative Non-Fiction. These responses originally generated by LLMs are further edited by 18 professional writers using the above-mentioned taxonomy, resulting in 8035 fine-grained edits (Section \ref{data_collection}). 

    \item We present a thorough analysis of the process of editing LLM-generated text, offering insights into how expert writers edit them, to what extent the edits differ in quantity, how the distribution of edit categories varies across text generated by different model families, and whether LLM generated text contain any specific stylistic idiosyncrasies (Sections \ref{data_collection} and \ref{style}).

    \item Building on prior work \citet{hayes1987cognitive,scardamalia1983development} we conduct an empirical investigation that tests if LLMs can automatically detect and rewrite their own idiosyncrasies. Our statistically significant results show an encouraging preference trend {$Writer-edited > LLM-edited > LLM-generated $} suggesting that edits improve human-AI alignment in the writing process. 

    \item Finally, we discuss how LLM edits can both mimic and differ from edits provided by professional writers, and what future LLM-based writing support tools can do to improve the co-writing experience.
\end{itemize}

Our code, data, and experimental setup is available at \footnote{\url{https://github.com/salesforce/creativity_eval/tree/main/Writing_Alignment}}.

\section{RELATED WORK}
\subsection{Text Editing in HCI}
Text editing is the process of modifying written content using specialized software. HCI research on text editing aims to improve digital writing tools' efficiency and usability. Word processors have long allowed flexible editing functions \cite{pea1987chapter}. Systems like Soylent \cite{10.1145/2791285}, MicroWriter \cite{10.1145/2858036.2858108}, and WearWrite \cite{10.1145/2858036.2858169} developed interfaces for crowd-based editing, focusing on task breakdown, cost management, and minimizing delays. \citet{10.1145/800045.801614} proposes a goal-fate analysis model for text editing behavior, supported by data showing distinct plan units in editing tasks, with the potential for intention-based user assistance. \citet{10.1145/800049.801803} investigated text editing skill progression and effective training methods. \citet{10.1145/800045.801604} examined the impact of experience on editing behavior, questioning if users naturally develop optimal strategies or plateau, noting that experienced users tend to develop more efficient editing heuristics than novices. \citet{10.1145/3613904.3641899} present ABScribe, a novel interface that streamlines the process of generating and comparing writing variations using Large Language Models, addressing challenges in existing text-editing workflows and improving writer's efficiency and satisfaction. \citet{10.1145/3635636.3656187} suggest that imperfect AI text suggestions can promote deeper engagement in rewriting, potentially preserving the writer's authenticity and creative ownership. \citet{10.1145/3544548.3581345} found that providing rationales for edits in collaborative writing was generally preferred by participants, despite no significant differences in survey results. This led to design recommendations for effective collaboration. \citet{10.1145/3526113.3545672} propose a text editor with continuously updated paragraph-wise summaries as margin annotations to help users plan, structure, and reflect on their writing process.\citet{laban2023beyond} introduce InkSync, an LLM-based editing interface suggesting executable document edits. It uses a three-stage approach (Warn, Verify, Audit) to reduce factual errors and enhance editing accuracy, efficiency, and user experience compared to standard chat interfaces. In contrast to existing research, we focus on edits as a method to improve human-AI alignment in writing assistance. Our work characterizes the undesirable aspects of AI writing informed by expert consensus and designs an approach to mitigate these through text editing.

\subsection{Text Editing in NLP}

NLP research has explored various text editing tasks \cite{laban-etal-2023-swipe,pryzant2020automatically,choi-etal-2021-decontextualization}. Adding to it the advent of Large Language Models has enabled AI-assisted writing tools \cite{buschek2021impact,ippolito-etal-2022-case}. \citet{faltings-etal-2021-text} release the WikiDocEdits dataset and propose an interactive text generation setting in which a user interacts with the system by issuing commands to edit existing text. \citet{raheja-etal-2023-coedit} proposed an instruction-based editing system using fine-tuned language models. \citet{shu2024rewritelm} developed strategies for cross-sentence rewriting and introduced the OpenRewriteEval benchmark. \citet{reid-neubig-2022-learning} modeled multi-step editing processes to better mimic human content creation and improve performance on various tasks. \citet{kim-etal-2022-improving} presented a system that iteratively improves fluency, clarity, coherence, and style by detecting editable spans and their corresponding edit intents, then instructing a revision model to refine the text. \citet{yang-etal-2017-identifying-semantic} developed a taxonomy and classifier for Wikipedia edit intentions. Following them \citet{du-etal-2022-understanding-iterative} created a multi-domain corpus of revised text with annotated edit intentions. Unlike existing work, we create a resource for text editing that caters to challenging writing tasks (literary fiction and creative nonfiction). Our data consists of 8035 fine-grained edits that are annotated by creative writing experts and we further show how recent advances in few-shot learning can help models improve their own writing by learning from edits provided by the writers.

\subsection{Issues in AI Writing}
Prior work has highlighted several issues in AI-generated text. \citet{10.1145/3613904.3642731, 10.1145/3635636.3656201, ippolito2022creative, mirowski2023cowriting,marco2024pron} show how LLM-generated text is often hackneyed and rife with clichés, lacks nuance, subtext, and rhetorical complexity. Recent work from \citet{mirowski2024robot} shows LLMs fail to act as good creativity support tools for comedy writing and mostly resort to producing bland and biased comedy tropes. They further highlight how existing moderation strategies used in safety filtering and instruction-tuned LLMs reinforce hegemonic viewpoints by erasing minority groups and their perspectives in writing. In summarizing short stories \citet{subbiah2024reading} demonstrate how LLMs struggle with specificity and interpretation of difficult subtext. In a similar vein, \citet{tian2024large} shed light on how LLM-generated stories are homogeneously positive and lack tension. Compared to existing work we create a fine-grained taxonomy highlighting the issues in AI writing and further create a large-scale corpus to fuel research in this direction. We also develop automated methods to identify and mitigate issues in AI writing at scale.

\subsection{Human AI alignment in Writing}

\citet{lee2024design} highlight how AI tools have transformed writing processes, establishing new criteria for future AI writing assistants. In a similar vein \citet{10.1145/3613904.3642625} reveal that while users benefit from AI assistance in productivity and confidence, potential drawbacks include decreased accountability and diversity in writing. LLMs used in writing assistance can significantly influence human-authored content. \citet{hohenstein2020ai} found LLM-generated text suggestions can affect a human writer's emotional tone. \citet{10.1145/3377325.3377523} showed predictive text encourages predictable writing. \citet{anderson2024homogenization} and \citet{laban2023beyond} found LLMs like ChatGPT helped users generate more detailed ideas, but outputs were less semantically distinct across users \cite{padmakumar2023does}, and participants felt less responsible for their produced ideas. Recent work from \citet{pan2024spontaneous} demonstrates language models can enhance outputs via feedback. However, methods like \textit{Iterative Self-refinement scenarios}, using another language model as an evaluator, may result in reward hacking, where the model exploits the evaluator's flaws. For alignment training, it's crucial to consider the absolute desirability of each potential response, not just how responses compare to one another in terms of preference. Towards this in our work, we create pairs consisting of an initial LLM-generated response and its refined counterparts that by nature are more contrastive (or closely comparable). Our results show that such a pairing results in improved alignment and agreement during preference ranking.

\section{DESIGN CONSIDERATIONS TO IMPROVE AI WRITING} \label{design}
``The secret to good writing is good editing.It's what separates hastily written, randomly punctuated, incoherent rants from learned polemics and op-eds, and cringe-worthy fan fiction from a critically acclaimed novel '' \cite{nytedit}. In this section, we outline the design principles and desiderata that guided our approach to improving AI writing through textual edits.

\paragraph{Design Principle 1: Develop a comprehensive edit taxonomy grounded in expert writing practices}

This principle emphasizes creating a comprehensive taxonomy of edit categories \cite{faigley1981analyzing} rooted in an expert editor and writer's practices. Prior work has shown that experts and novices define revising in very different ways with experts attending more systematically to different aspects of the text than novices \cite{bracewell1978development,sommers1980revision,10.1145/800049.801803,10.1145/800045.801604}. By developing such a taxonomy, we aim to provide an approach to analyzing and enhancing LLM-generated text. It also allows for a more granular understanding of the specific areas where AI writing may fall short and enables targeted improvements. \citet{sommers1980revision} found that ``experienced writers have a second objective; a concern for their readership''. Grounding the taxonomy in expert writing practices ensures that the edits align with the standards of high-quality writing and are acceptable to its readers. Finally, this principle also acknowledges the complexity of the editing process, recognizing that different categories of edits may be required at various levels of the text, from sentence-level corrections to broader structural changes \cite{hayes1987cognitive, sommers1980revision,yang-etal-2017-identifying-semantic}. 

\paragraph{Design Principle 2: Leverage edits to balance both meaning preservation and substantive semantic changes}
Preserving the core meaning and intent of the original text is crucial to maintaining coherence and faithfulness to the initial ideas. On the other hand, introducing substantive semantic changes is often required to adhere to the quality and characteristics of good writing. Prior work on edit taxonomies focuses on low-level syntactic operations \cite{faigley1981analyzing} or semantic edits \cite{yang-etal-2017-identifying-semantic,daxenberger-gurevych-2013-automatically,laban-etal-2023-swipe} tailored to specific websites like Wikipedia. LLM-generated text often benefits from syntactic edits. These edits (primarily meaning preserving) enhance readability by diversifying sentence structures, expanding vocabulary choices, and minimizing repetitive phrasing. Consequently, semantic edits (both meaning preserving and changing) in AI writing are important for enhancing specificity or reducing unnecessary flourishes and clichés that can otherwise obscure meaning. Our methodology aims to navigate the tension between maintaining original meaning and introducing necessary improvements to mitigate AI-specific writing quirks.

\paragraph{Design Principle 3: Utilize edits as a mechanism for enhancing human-AI alignment in writing}
Current AI writing systems are developed using pre-trained language models (LMs) refined through human interaction, employing supervised learning and reinforcement learning (RL) techniques. Reinforcement learning from human feedback (RLHF) \cite{ziegler2019fine} is a key approach, transforming human input into training data to guide LMs toward desired outcomes. The most common type of feedback used with RLHF is binary preferences between pairs of examples sampled from one or more Language Models \cite{christiano2017deep}. However, a learned preference ordering can fail to converge to the true one when the desirability of examples depends on noise \cite{gao2024impact}. Following recent work in preference learning \cite{dubey2024llama,d2024anchored}, we evaluate edits as a mechanism for enhancing alignment. An LLM-generated response that has been edited typically contains fewer undesirable traits and can be paired with the original LLM-generated response for preference ranking ($edited > original$).While contemporaneous works \cite{dubey2024llama, d2024anchored} have conducted preliminary efforts to incorporate edits for improved preference data collection, we evaluate this approach in the context of creative writing.

\section{LARGE SCALE DATA COLLECTION PROCESS} \label{data_collection}
\begin{figure*}
    \centering
    \includegraphics[width=\textwidth]{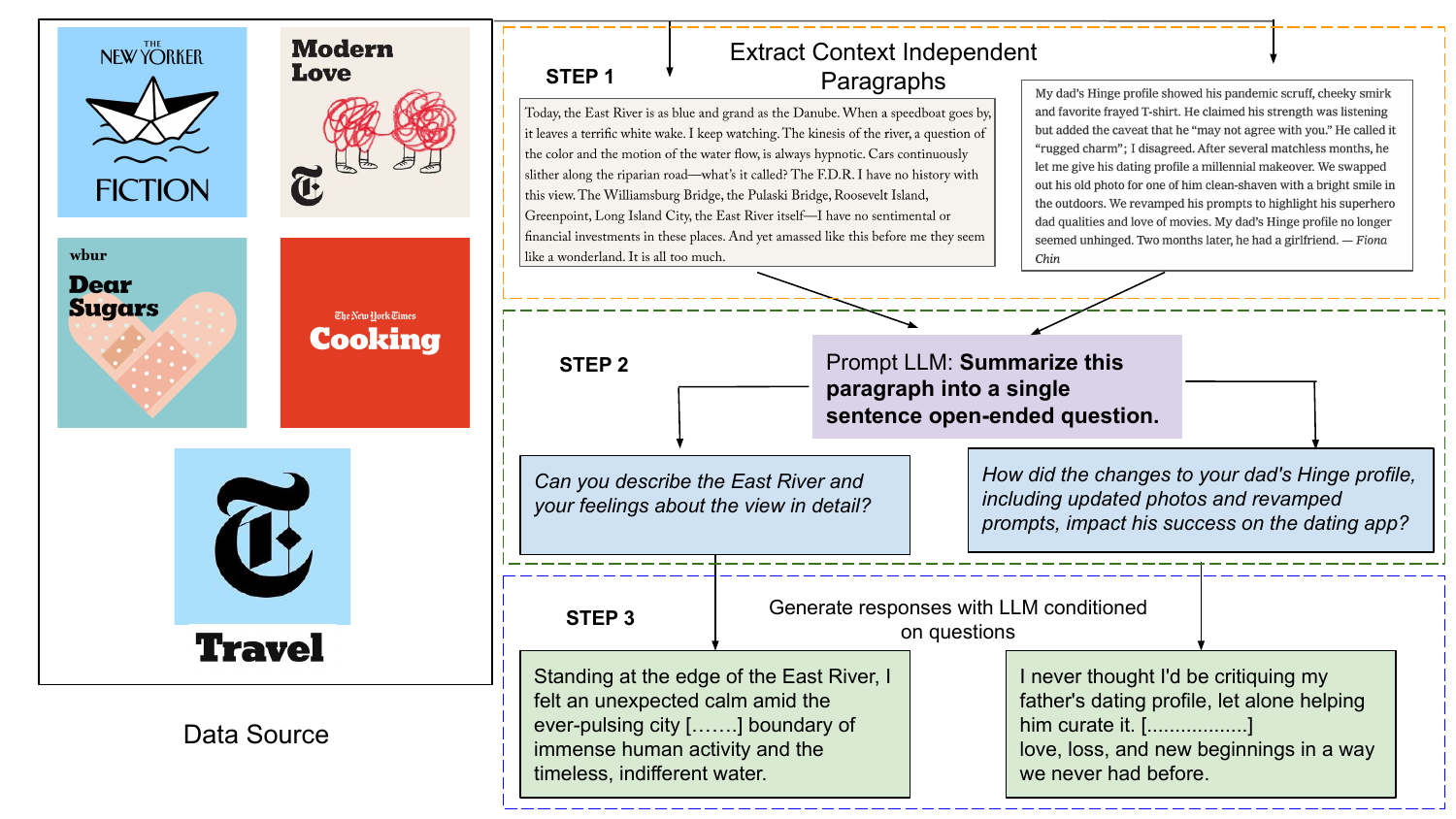}
    \caption{\label{data_pipeline} The pipeline for data creation. Step 1) Extracting context-independent paragraphs from our respective sources Step 2) Using an LLM to automatically generate instructions for corresponding human-written text Step 3) Use the generated instructions grounded in real-world writing to elicit responses from LLMs to create $<$instructions,response$>$ pairs}
\end{figure*}

We aim to create a valid test-bed to evaluate the quality of LLM-generated text on realistic writing tasks that require creative skill. We follow a three-step approach illustrated in Figure~\ref{data_pipeline}: (1) First we select original paragraphs of human-written text from trusted publication venues, (2) Second we reverse-engineer each of these paragraphs into a writing instruction. Because each instruction originates from an existing human-written paragraph within a piece of creative writing, this simulates real-world writing situations. (3) Third, we prompt several LLMs to generate responses to each of the writing instructions. In the following subsections, we first detail each of these steps, and then describe the formative study we conducted to develop a taxonomy of idiosyncrasies in LLM-generated text.

\subsection{Collecting Instruction and Response Pairs}

\begin{table*}[!ht]
\small
\renewcommand{\arraystretch}{1.2}
\begin{tabular}{|l|p{1.8cm}|l|l|}
\hline
Domain & Source \& Genre & Example Seed Paragraph with Generated Instruction & \#N \\ \hline
Fiction & \begin{tabular}[c]{@{}l@{}}\href{https://www.newyorker.com/fiction-and-poetry}{The NewYorker}\\ (Literary Fiction)\end{tabular} & \begin{tabular}[c]{@{}l@{}}The sunset is a red-gold rumpus on the western sky. It has rained. The crow tosses \\itself from branch to branch, pole to pole, glistening on its pace, and she follows. They\\ are soon far from where they began, streets unfamiliar to her, an older part of town,[.....\\....]. A man reading. Old Christmas tree in a corner. It  feels secret. The sky is \\clearing overhead. She feels secret, too. She feels tremendous.\vspace{1ex}\\
{\color{blue}Instruction} : \textbf{Can you describe a vivid scene at sunset that transitions into}\\\textbf{nighttime, incorporating elements of nature, urban surroundings, and} \\ \textbf{personal observations?}\end{tabular} & 815   \\ \hline\hline
\multirow{4}{*}{\begin{tabular}[c]{@{}l@{}}Creative \\Non-Fiction\end{tabular}} & \begin{tabular}[c]{@{}l@{}}\href{https://www.nytimes.com/column/36-hours}{NYTimes}\\ (Travel Writing)\end{tabular}         & \begin{tabular}[c]{@{}l@{}} Prague, the Czech capital, is finding a new balance between preserving its past and\\ embracing the future, improving many of its important historic sites while making \\striking additions to its skyline. [.........] Stop by for a coffee, hit up one of the many\\ great new bakeries or visit a charismatic old beer hall as you explore a city that is \\clearly entering its prime. \vspace{1ex}\\ 
{\color{blue}Instruction} : \textbf{How is Prague balancing its historical preservation with modern}\\\textbf{development while enhancing local amenities and vibrant} \\ \textbf{neighborhoods outside Old Town?}

\end{tabular} & 110   \\ \cline{2-4}
                                      & \begin{tabular}[c]{@{}l@{}}\href{https://cooking.nytimes.com/recipes/}{NYTimes}\\ (Food Writing)\end{tabular} & \begin{tabular}[c]{@{}l@{}}The origins of the fruit sandwich are believed to go back to Japan’s luxury fruit stores \& \\ the fruit parlors attached to them. This version comes from Yudai Kanayama, a native of \\ Hokkaido who runs the restaurants the Izakaya NYC and Dr Clark in New York. [.........]\\ .The sandwich looks like dessert but isn’t, or not exactly; it makes for a lovely little meal \\that feels slightly illicit, as if for a moment there are no rules
                                      \vspace{1ex}\\
                                      {\color{blue}Instruction} : \textbf{How did Yudai Kanayama reinvent the traditional Japanese fruit}\\\textbf{
sandwich to create a unique culinary experience?}

                                      \end{tabular} & 83    \\ \cline{2-4} 
                                      & \begin{tabular}[c]{@{}l@{}}\href{https://www.nytimes.com/column/modern-love}{NYTimes}\\ (Personal Essay)\end{tabular}         & \begin{tabular}[c]{@{}l@{}}My dad’s Hinge profile showed his pandemic scruff, cheeky smirk and favorite frayed \\ T-shirt. He claimed his strength was listening [.........] We revamped his prompts to \\highlight his superhero dad qualities and love of movies. My dad’s Hinge profile no \\longer seemed unhinged. Two months later, he had a girlfriend.\vspace{1ex}\\
                                       {\color{blue}Instruction} : \textbf{How did the changes to your dad’s Hinge profile, including updated}\\\textbf{
photos and revamped prompts, impact his success on the dating app?}

                                      \end{tabular}                                                                                                                                                                                                            & 19    \\ \cline{2-4} 
                                      & \begin{tabular}[c]{@{}l@{}}\href{https://therumpus.net/sections/blogs/dear-sugar/}{Dear Sugar}\\ (Internet Advice)\end{tabular}     & \begin{tabular}[c]{@{}l@{}}What is a prestigious college? What did attending such a school allow you to believe \\about yourself? What assumptions do you have about the colleges that you would not \\describe as prestigious? What sorts of people go to prestigious colleges and not [.........]\\ I believe our early experiences and beliefs about our place in the world inform who we \\think we are and what we deserve and by what means it should be given to us. \vspace{1ex}\\
                                      {\color{blue}Instruction} : \textbf{How do your beliefs and assumptions about educational privilege}\\\textbf{
and the type of schools people attend shape your current view of yourself}\\ \textbf{and others?}
                                      \end{tabular}                                                                                                                                                                                                                                       & 30    \\ \hline
\end{tabular}
\vspace{2ex}
\caption{\label{table:source_paragraph}Venues of source material used to extract real-world examples of creative writing, along with example seed paragraphs and the generated instructions and the number of samples per genre.}
\end{table*}

To curate source material, we select five well-regarded \footnote{\url{https://www.pewresearch.org/politics/2012/09/27/section-4-demographics-and-political-views-of-news-audiences/}} publication venues, listed in Table \ref{table:source_paragraph}, that publish pieces in different domains ranging from fiction to food writing and internet advice. For each venue, we manually extract between 100-700 pieces of writing and isolate individual paragraphs. We then manually review these paragraphs, ensuring they are long enough and can stand alone as coherent pieces of writing without requiring additional context. In total, we selected approximately 1200 paragraphs following this procedure. The Literary Fiction genre has a larger representation (80\%) in our selection, while the creative non-fiction genres have a smaller representation.

Next, we follow \citet{li2023self}'s approach of \textit{Instruction Backtranslation} to automatically generate instructions corresponding to each of the selected paragraphs. Specifically, we prompt an LLM (i.e., GPT4o) to summarize each paragraph into an open-ended question. Questions obtained through back-translation (see examples in Table~\ref{table:source_paragraph}) can be interpreted as realistic writing instructions. We manually verified the generated instructions, filtering out questions that were ill-formed or overly specific, yielding a total of 1,057 writing instructions.

Finally, we use the generated instructions to elicit responses from three state-of-the-art LLMs: OpenAI's GPT-4o \cite{OpenAI2023GPT4TR}, Anthropic's Claude-3.5-Sonnet \cite{Claude3.5Sonnet}, and Meta's Llama 3.1-70b \cite{dubey2024llama}. Each LLM is used to generate responses to one-third of our instruction data. We ensure that each LLM responds to instruction across all domains in equal proportion. To generate high-quality responses, we provide each LLM with the writing instruction, as well as the genre and source, and instruct it to adhere to the style of the venue. The prompt further specifies: \textit{``Try your best to be original, avoiding clichés or overused tropes. Do not use ornamental language and focus on nuance, simplicity, and subtext''} (See Prompt in Appendix \ref{sec:appendix} Table \ref{prompts}). Through this process, we obtain 1,057 writing $<$instructions, response$>$ pairs, with responses averaging 205 words. This collection of instructions and LLM-generated responses serves as the foundation for the three studies we conducted: the formative study, the full-scale editing annotation, and the preference annotation.

\subsection{Formative study: formulating the taxonomy for fine-grained edits}

Our formative study observed writers with copy-editing experience as they edited LLM-generated text in the creative writing domain. We aimed to identify common edit categories. The study consisted of three phases. First, participants were individually briefed via video conference on the study's objectives. Next, they accessed a web application (Figure \ref{pilot}) to view $<$instruction, response$>$ pairs from our dataset (Section \ref{data}). For each sample, participants highlighted problematic response spans, suggested rewrites, and tagged each span with a free-form category to characterize the issue. We recruited eight participants for the formative study, with each completing annotations for 25 samples.

\begin{table*}[!ht]
    \centering
    \small
    \renewcommand{\arraystretch}{1.05}
    \begin{tabular}{|l|l|l|l|l|}
    \hline
    ID & Profession & Gender & Age & Educational Background \\ \hline
    W1 & Writer \& Editor at Magazine & Male & 28 & MFA in Fiction \\ \hline
    W2 & Writer \& Fellow at Fine Arts Work Center & Male & 29 & MFA in Fiction\\ \hline
    W3 & MFA Fiction Student & Male & 31 & MFA in Fiction\\ \hline
    W4 & Writer & Female & 30 & MFA in Non-Fiction \\ \hline
    W5 & Writer & Female & 35 & MFA in Fiction\\ \hline
    W6 & MFA Poetry Student & Female & 27 & MFA in Poetry\\ \hline
    W7 & Writer \& Journalist & Female & 35 & MFA in Non-Fiction\\ \hline
    W8 & MFA Fiction Student & Male & 26 & MFA in Fiction\\ \hline
    \end{tabular}
    \vspace{1ex}
    \caption{\label{participants} Pilot study: background of participants.}
    \vspace{-5ex}
\end{table*}

For participant selection in our formative study, we limited involvement to individuals with established expertise in creative writing. Participants were required to have completed a Master of Fine Arts (MFA) in Creative Writing and were recruited through mailing lists from MFA writing programs in the United States. This aligns with prior work from \citet{10.1145/3613904.3642731}, using the Consensual Assessment Technique \cite{amabile1982social}, which emphasizes the importance of recruiting domain experts. During the initial video call, we confirmed participants' familiarity with copy-editing and informed them they would edit LLM-generated texts. We recruited participants through UserInterviews~\footnote{\url{https://www.userinterviews.com}}, a professional freelancing platform, paying \$75 USD for study completion. Editing a response took 4-6 minutes, with all participants finishing within two hours. Table \ref{participants} shows diverse demographics and professional backgrounds of the recruited creative writers.\footnote{The research was conducted at an institution without a formal IRB approval process. However, an Ethical Practices team reviewed the work and study protocols. No personally identifiable information (PII) was collected or shared during data collection, and participants were offered compensation regardless of study completion.} In total, the eight participants edited 200 paragraphs, annotating roughly 1,600 edits attributed to 50 distinct initial categories. We used this data as the foundation for our next analysis, which aimed to develop a taxonomy for categorizing edits.

\subsection{From initial to final categorization of edits}

\begin{table*}[!ht]
\centering
\small
\renewcommand{\arraystretch}{1.35}
\begin{tabular}{|l|l|l|}
\hline
Final Category & Initial Categories & Participants \\ \hline
Cliche & \textit{Cliched image from old westerns, Cliche, Hackneyed} & \begin{tabular}[c]{@{}l@{}}W1,W2,W3,W4, \\ W5,W6,W7,W8\end{tabular} \\ \hline\hline
\begin{tabular}[c]{@{}l@{}}Unnecessary/Redundant \\ Exposition\end{tabular} & \begin{tabular}[c]{@{}l@{}}\textit{Repetition of what has already been stated, Unnecessary, Show don't tell,}\\ \textit{Repetition, Cut Unnecessary, Unnecessary because implied, Over } \\ \textit{exposition, Fluff, Slim down, Trying to cut things down, Concision}\end{tabular} & \begin{tabular}[c]{@{}l@{}}W1,W2,W3,W4,\\  W5,W6,W7,W8\end{tabular} \\ \hline\hline
Purple Prose & \begin{tabular}[c]{@{}l@{}} \textit{Too wordy, Purple Prose, Ornamental, Very Verbose, Clunky}\\ \textit{Unnecessarily wordy, Simplify, Overwrought, Mixed metaphor}\end{tabular} & \begin{tabular}[c]{@{}l@{}}W1,W3, W4\\ W5,W7,W8\end{tabular} \\ \hline\hline
\begin{tabular}[c]{@{}l@{}}Poor Sentence Structure\end{tabular} & \begin{tabular}[c]{@{}l@{}}\textit{Structure, Transition, Editing for clarity, Better to split up into two}\\ \textit{sentences, Run-on sentence, Very Long and Complex Sentence}\end{tabular} & W1, W5, W7,W8 \\ \hline\hline
\begin{tabular}[c]{@{}l@{}}Lack of Specificity\\ and Detail\end{tabular}     & \begin{tabular}[c]{@{}l@{}}\textit{Lacks specificity, Overly General, More details to help move the reader,} \\ \textit{Added details, Creating a scene, Contextualizing information},\\ \textit{Deepening internality, Needs to be more specific, Adding Voice}\end{tabular} & \begin{tabular}[c]{@{}l@{}}W1,W2,W3,W4\\ W5,W6,W8\end{tabular}      \\ \hline\hline
\begin{tabular}[c]{@{}l@{}}Awkward Word Choice\\ and Phrasing\end{tabular}    & \begin{tabular}[c]{@{}l@{}}\textit{Word Choice, Pronoun Clarity, Passive, Awkward Word Choice,}\\ \textit{Wrong choice of word, Rewording, Rephrasing, Weird Phrasing}, \\ \textit{Inelegant}\end{tabular} & \begin{tabular}[c]{@{}l@{}}W1,W2,W3,W4\\ W5,W6,W7,W8\end{tabular}   \\ \hline\hline
\begin{tabular}[c]{@{}l@{}}Tense Inconsistency\end{tabular} & \begin{tabular}[c]{@{}l@{}}\textit{Fragment sounds weird-is it past or present?, Wrong Tense}, \\ \textit{Inconsistent  Tense}\end{tabular} & W1, W5, W7,W8                                                       \\ \hline
\end{tabular}
\vspace{1ex}
\caption{\label{taxonomy}Our final taxonomy for fine-grained edits to mitigate idiosyncrasies in AI writing}
\vspace{-5ex}
\end{table*}

We observed significant semantic overlap among the 50 initial categories used by participants, suggesting potential consolidation into a unified taxonomy. For instance, ``Show don't tell'' (W4) corresponded to ``Unnecessary because implied'' (W6). Using a general inductive approach for qualitative data analysis \cite{thomas2006general}, we synthesized these 50 initial categories into a comprehensive, fine-grained taxonomy of edits. First, two authors independently bucketed these categories into initial low-level groups. Through iterative discussions, these groups were refined to reduce overlap and establish shared groupings. The refined low-level groups were then aggregated into high-level categories. Each high-level category was assigned a name reflecting its generalized representation.

The aggregation process yielded 7 distinct edit categories, presented in Table~\ref{taxonomy} along with contributing participant IDs. Final categories were retained only if derived from initial categories identified by at least four participants, ensuring majority representation in editing feedback \footnote{Only 5\% of edit categories were not included in the 7 categories as they did not have enough coverage}. It's worth noting that not every LLM-generated response exhibits all these idiosyncrasies. The formative study's objective was not to establish the relative prevalence of each category. Instead, this taxonomy serves as a useful framework when considering the categories of edits to apply to LLM-generated content. The categorization provides a structured approach to refining such text.  

\subsection{Final Taxonomy for Fine-Grained Edits} \label{fine_grained_taxonomy}
Here we describe our final taxonomy for fine-grained edits. Table \ref{example_edits} shows examples of edits in each of these categories defined below
\subsubsection{\textbf{Cliché}}
Clichés in writing are pejoratively characterized as phrases, ideas, or sentences overused to the point of losing their original impact or meaning. They often use vivid analogies or exaggerations from everyday life to describe abstract concepts. While occasionally effective when used sparingly, the frequent use of clichés in writing is generally viewed as a sign of inexperience or lack of originality \cite{fountain2012clichés}. Replacing clichés with fresh, original language improves the writing and engages readers more effectively.
\subsubsection{\textbf{Unnecessary/Redundant Exposition}}
Unnecessary or redundant exposition refers to the inclusion of excessive, repetitive, or implied information in writing. This common pitfall often involves restating the obvious or providing details that add little value. In a conversation with W2 they said \textit{``I'm adding a category of edit called ``fluff" - this is a common term in the writing world to refer to unnecessary filler"}. Effective writing embraces the principle of ``show, don't tell,'' allowing readers to infer meaning from context rather than relying on explicit explanations \cite{noble2013show,chekhov1984selected,king2000writing,burroway2019writing}. Impactful writing, often allows the core message to shine through without being obscured by unnecessary verbiage.
\subsubsection{\textbf{Purple Prose}} In literary criticism, purple prose refers to excessively elaborate writing that disrupts the narrative flow by attracting undue attention to its flamboyant style \cite{wikipedia_purple_prose}. This can detract from the text's overall appreciation. Such writing is often difficult to read, using sprawling sentences, abstract words, and excessive adjectives, adverbs, and metaphors to convey little information. Careful editing can trim purple prose by replacing ornate language with more direct expressions, resulting in clearer writing that preserves narrative flow and the author's voice.
\subsubsection{\textbf{Poor Sentence Structure}} Poor sentence structure reduces the clarity and readability of writing \cite{bormuth1966readability,klare1974assessing,meyer2003text}. A lack of proper transitions can make the text feel disjointed and hard to follow. Editing for clarity \cite{dermer2009fluency} often reveals that it's better to split a convoluted thought into two sentences, rather than forcing it into one \cite{lane1999writing}. Run-on sentences, characterized by multiple independent clauses improperly connected, are also frequent problems in AI writing \cite{10.1145/3613904.3642731}. These, very long and complex sentences can overwhelm the reader, making the core message difficult to grasp. Edits that reduce these problems lead to more coherent and fluent text.
\subsubsection{\textbf{Lack of Specificity and Detail}} Lack of specificity and details in writing often stems from a writer's tendency to rely on broad generalizations \cite{macdonald1986specificity}. This overly general approach fails to engage readers, leaving them unable to visualize scenes or connect with any given writing on a deeper level. Good writing often focuses on adding vivid details that create a clear mental image \cite{doty2014art,fleckenstein1991inner,kuzmivcova2014literary}, contextualizing information to give it relevance \cite{rababah2022contextualization}, and deepening the internality of characters or subjects \cite{currie1990nature,fishelov1990types}. Additionally, developing a unique voice through carefully chosen words and phrases can inject personality into the writing, making it more engaging and distinctive \cite{nelson2012academic,herring2007writing}. Edits belonging to this category typically make the text longer as writers add more details to make the text engaging.
\subsubsection{\textbf{Awkward Word Choice and Phrasing}} Awkward phrasing can significantly reduce writing quality, often confusing or disengaging readers. This issue typically involves misused or disproportionate use of certain words \cite{kobak2024delving}, unclear pronoun references, or an overuse of passive voice. In an email, W1 pointed out ``\textit{Another little observation to share: a very common phrasing in these excerpts is `seem to \_(verb)\_'. This is not technically wrong it's just inelegant, something many writing teachers have told me to avoid. Unless there is some specific uncertainty or doubt about the verb action, it's always preferred to just use the verb without `seem' (ex. from the current excerpt I have up: `amplified' is better than `seemed to amplify')}". Editing plays a crucial role in refining these elements. Through careful revision, writers can identify and replace imprecise or ill-fitting words with more appropriate alternatives, ensuring each term accurately conveys the intended meaning.
\subsubsection{\textbf{Tense Inconsistency}} Tense inconsistency is a prevalent issue in writing. It occurs when a writer inadvertently shifts between past, present, and future tenses often even within the same paragraph or sentence. This grammatical misstep can make the timeline of events unclear and detract from the overall coherence of the text. Careful editing plays a crucial role in addressing this issue. By paying close attention to verb forms and temporal indicators, editors can improve writing that deals with tense inconsistency \cite{moens1987tense}.

\begin{figure*}
\footnotesize
\stackunder[5pt]{\frame{\includegraphics[width=0.9\textwidth]{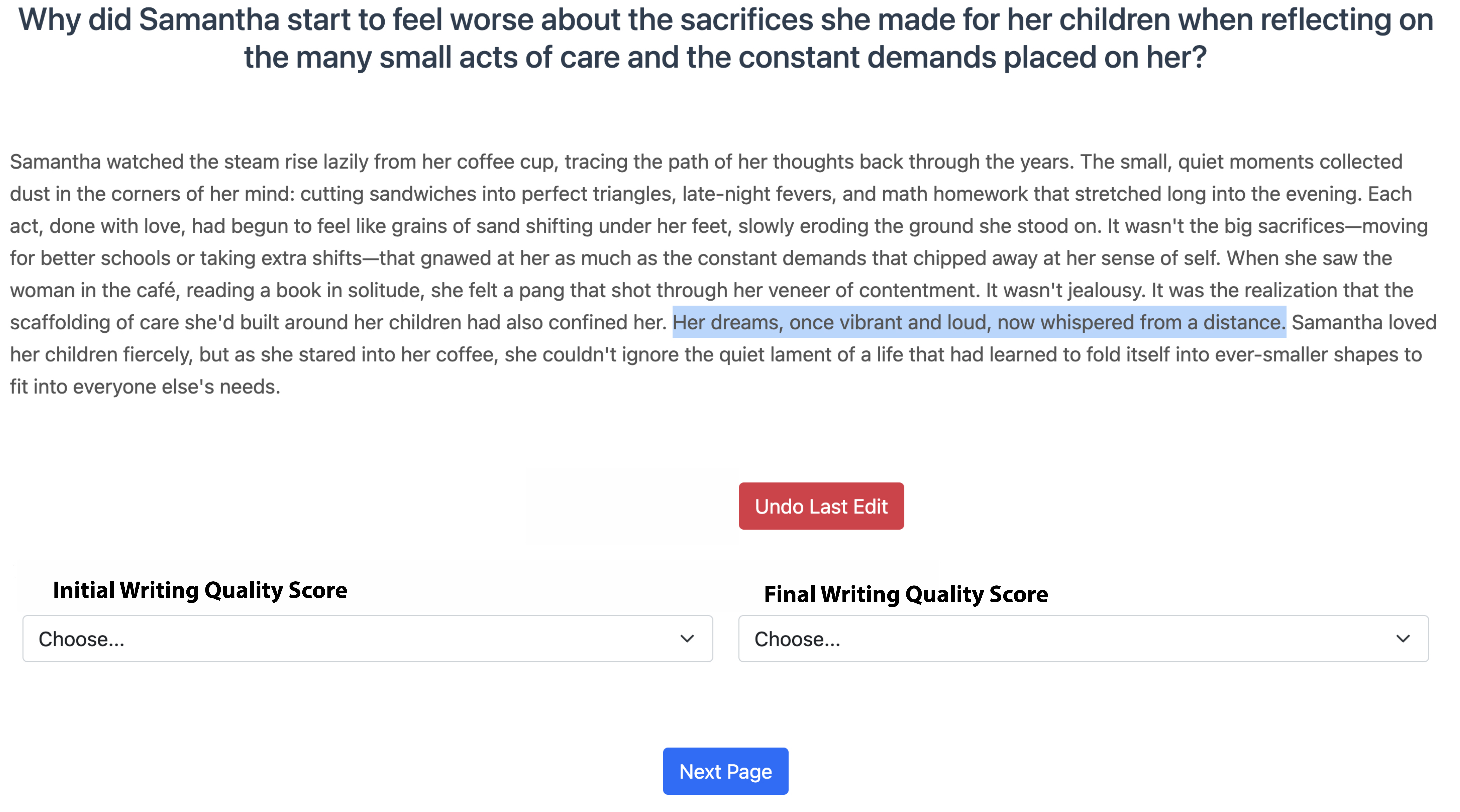}}}{Interface to collect edits on LLM-generated responses. Participants can click on any span they want to edit}%
\hspace{1ex}%
\stackunder[5pt]{\frame{\includegraphics[width=0.9\textwidth]{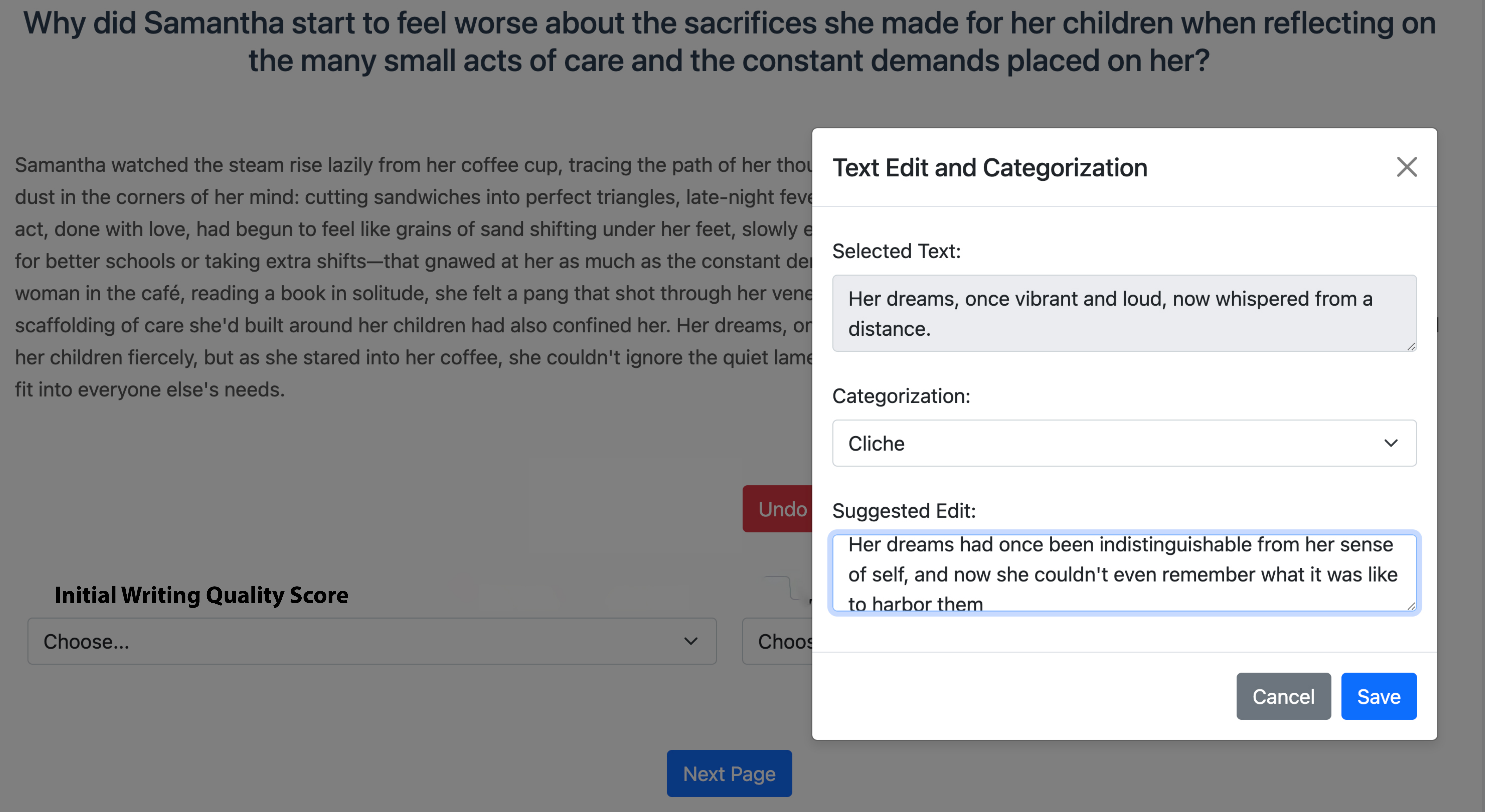}}}{Pop-up edit window to input edit and label the category of edit}
\caption{\label{interface} Interface to collect edits from writers on LLM-generated text}\end{figure*}

\begin{table*}[!ht]
\centering
\small
\begin{tabular}{|l|l|l|l|l|l|l|}
\hline
ID & Profession & Gender & Age & Educational Background& Responses\_Edited  \\ \hline
W1 & Writer \& Editor at Magazine & Male & 28 & MFA in Fiction & 123\\ \hline
W2 & Writer \& Fellow at Fine Arts Work Center & Male & 29 & MFA in Fiction &109 \\ \hline
W3 & Writer \& Teacher & Male & 31 & MFA in Fiction & 119\\ \hline
W4 & MFA Fiction Student \& Translator & Male & 30 & MFA in Fiction & 25 \\ \hline
W5 & Writer & Female & 35 & MFA in Poetry &77 \\ \hline
W6 & MFA Poetry Student & Female & 27 & MFA in Poetry & 23  \\ \hline
W7 & Writer \& Journalist \& MFA Fiction Student & Female & 35 & MFA in Fiction & 71 \\ \hline
W8 & MFA Fiction Student & Male & 26 & MFA in Fiction &23\\ \hline
W9 & Writer \& Editor & Male & 30 & MFA in Fiction & 25\\ \hline
W10 & Writer \& Creative Writing Instructor & Female & 28 & MFA in Fiction & 25\\ \hline
W11 & Writer & Female & 27 & MFA in Poetry & 25 \\ \hline
W12 & Writer \& High School Teacher & Male & 33 & MFA in Fiction & 24 \\ \hline
W13 & Writer \& Editor & Female & 29 & MFA in Non-Fiction & 25\\ \hline
W14 & MFA Fiction Student & Male & 26 & MFA in Fiction & 24\\ \hline
W15 & Poet & Female & 28 & MFA in Poetry & 125 \\ \hline
W16 & Writer \& Director & Male & 31 & MFA in Literary Arts& 122 \\ \hline
W17 & Writer & Non-Binary & 28 & MFA in Poetry & 25 \\ \hline
W18 & Screenwriter \& MFA Literary Arts Student & Female & 27 & MFA in Literary Arts & 67\\ \hline
\end{tabular}
\vspace{2ex}
\caption{\label{edit_participants}Background of participants who provide span level edits on LLM generated responses}
\end{table*}

\begin{figure*}
    \centering
    \begin{minipage}[t]{.45\textwidth}
        \subfloat[Edit Operations / Writer]
        {\label{fig:data_plot1} \includegraphics[width=\textwidth]{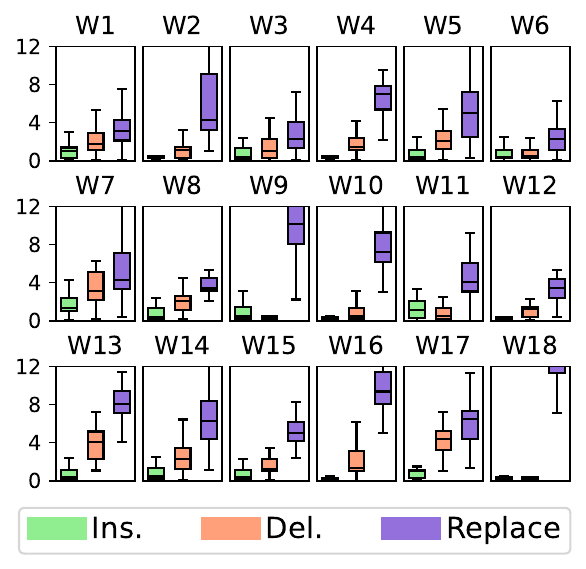}}
    \end{minipage}
    \begin{minipage}[t]{0.45\textwidth}
        \subfloat[Writing Quality Scores / Writer]
        {\label{fig:data_plot2} \includegraphics[width=\textwidth]{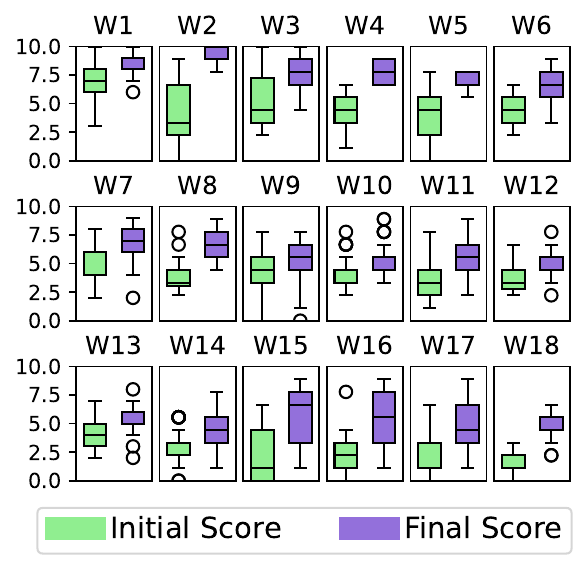}}
    \end{minipage}
    \begin{minipage}[t]{.25\textwidth}
        \subfloat[IWQS / Model]
        {\label{fig:data_plot3} \includegraphics[width=\textwidth]{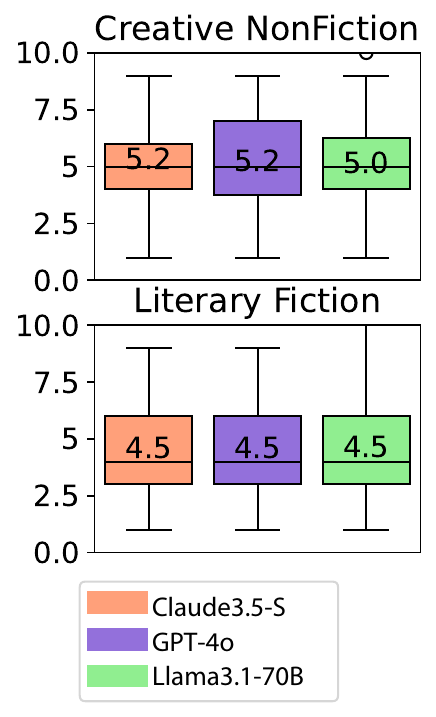}}
    \end{minipage}
    \begin{minipage}[t]{.4\textwidth}
        \subfloat[Editing Amount / IWQS]
        {\label{fig:data_plot4} \includegraphics[width=\textwidth]{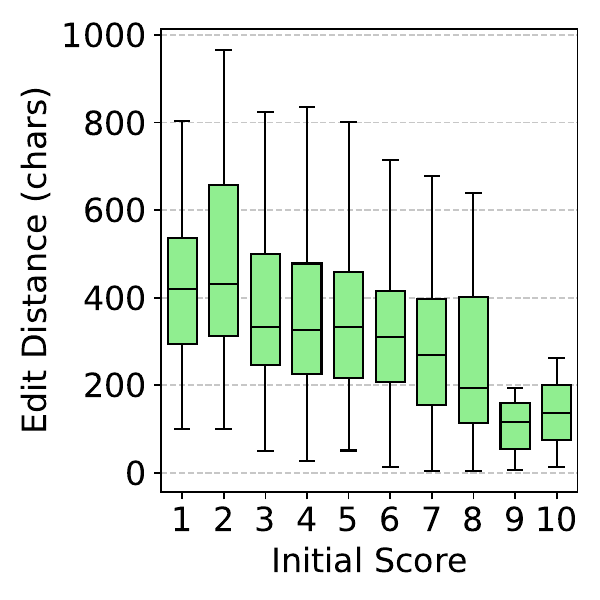}}
    \end{minipage}
    \hfill
    \caption{Analysis of 1,057 paragraphs edited by 18 Writer participants, analyzing: (\subref{fig:data_plot1}) the edit operations they perform (insertions, deletions, etc.), (\subref{fig:data_plot2}) the writing quality scores they assign, (\subref{fig:data_plot3}) comparing writing quality scores across LLMs, (\subref{fig:data_plot4}) the relationship between IWQS and editing amount.}
    \label{fig:data_analysis1}
    \vspace{-15pt}
\end{figure*}

\begin{figure*}
    \centering
 \begin{minipage}[t]{.3\textwidth}
        \subfloat[Edit Categories / Model]
        {\label{fig:data_plot5} \includegraphics[width=\textwidth]{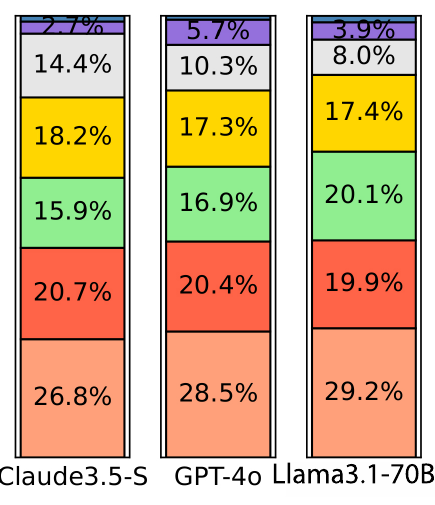}}
    \end{minipage}
    \begin{minipage}[t]{.65\textwidth}
        \subfloat[Edit Categories / Score]
        {\label{fig:data_plot6} \includegraphics[width=\textwidth]{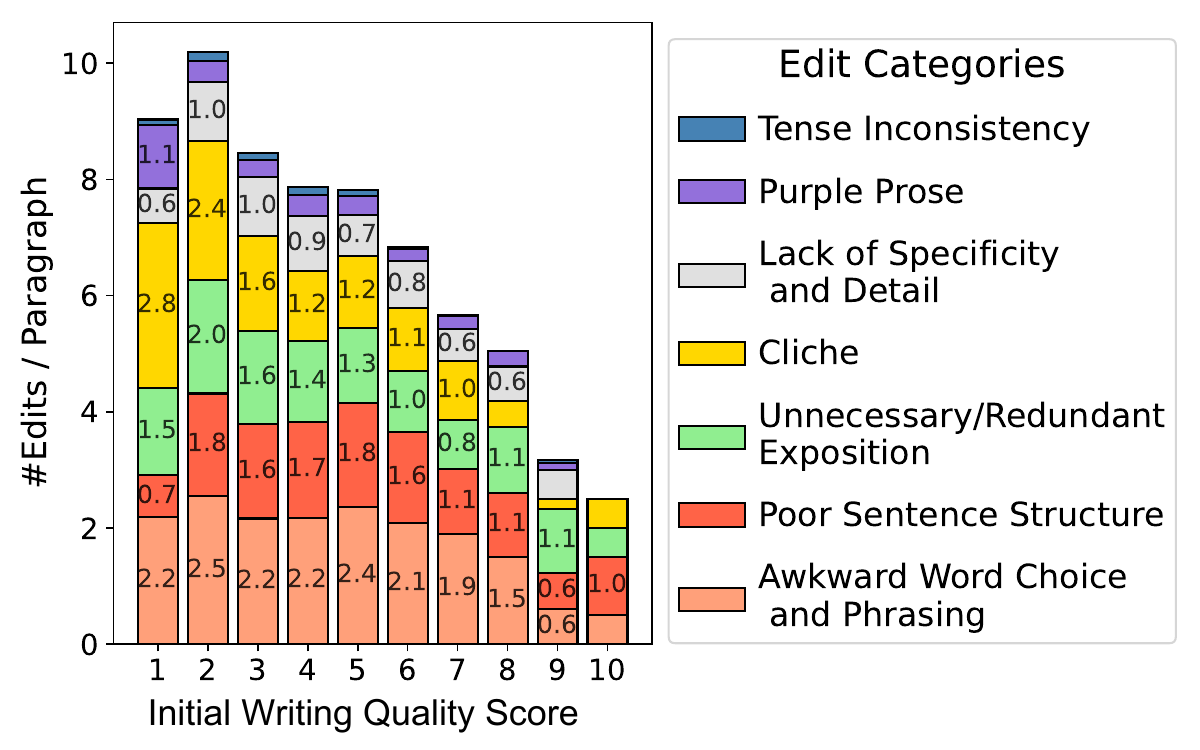}}
    \end{minipage}
    \hfill
        \begin{minipage}[t]{.65\textwidth}
        \subfloat[Distribution of semantic similarity scores between original and edited span]
        {\label{fig:data_plot7} \includegraphics[width=\textwidth]{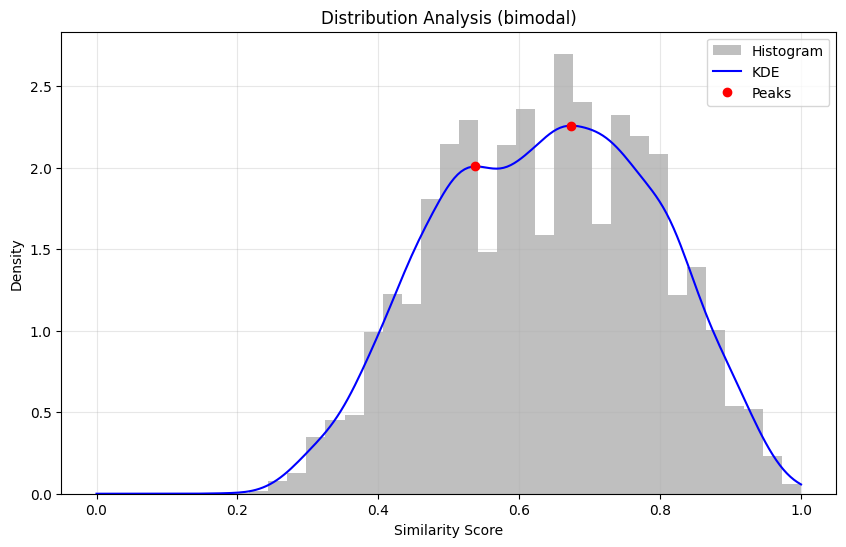}}
    \end{minipage}
    \hfill
    \caption{(\subref{fig:data_plot5}) the categories of edits they implement, and (\subref{fig:data_plot6}) the relationship between writing quality scores and error categories. (\subref{fig:data_plot7}) Distribution of semantic similarity scores for the edits in the dataset}
 \label{fig:data_analysis2}
    \vspace{-2pt}
\end{figure*}

\subsection{Collecting Edits on LLM Generated responses}
\begin{table*}[!ht]
\centering
\small
\renewcommand{\arraystretch}{1.15}
\begin{tabular}{|l|l|}
\hline
Categorization & Paragraph with rewrites \\ \hline
Cliche & \begin{tabular}[c]{@{}l@{}}As Sarah stepped off the bus, the scent of pine and damp earth enveloped her. {[}......{]}  In the kitchen, she found \\ herself reaching for the cabinet where her mother always kept the coffee, only to stop short. \textbf{\color{red}\st{The realization}} \\ \textbf{\color{red}\st{that she was alone here, truly alone, settled over her like a heavy blanket.}} \textbf{\color{ForestGreen}This time, though, she was}\\ \textbf{\color{ForestGreen}alone. Her mother would never come back.}She sank into a chair at the old oak table{[}....{]}\end{tabular} \\ \hline\hline
\begin{tabular}[c]{@{}l@{}}Unnecessary\\ / Redundant\\ Exposition\end{tabular} & \begin{tabular}[c]{@{}l@{}}As Mingus and Dylan stepped out of the car, {[}...{]} The Brooklyn-Queens Expressway \textbf{\color{red}\st{loomed above, a concrete}} \\ \textbf{\color{red}\st{behemoth that cast long shadows over the desolate landscape.}} \textbf{\color{ForestGreen}cast a long shadow.}{[}...{]} For a moment, he \\stood there, lost in thought, as the city seemed to hold its breath around him.\end{tabular}                                                                                                           \\ \hline\hline                        \begin{tabular}[c]{@{}l@{}}Lack of \\Specificity \\and Detail\end{tabular}                                                     &  \begin{tabular}[c]{@{}l@{}} \textbf{\color{red}\st{Dr. Arthur Steiger's fall from grace began with a series of whispered concerns among his colleagues}} \\\textbf{\color{red}\st{at Cormac General Hospital.}} \textbf{\color{ForestGreen}Pain was Dr. Arthur Steiger's forte. Not inflicting it, that is, but resolving }\\\textbf{\color{ForestGreen}it. Whenever a patient had problem, whether a tear in a tendon, a sprain, a knock, a headache, a}\\\textbf{\color{ForestGreen}broken bone – it was Dr. Steiger that knew what to do.}The small-town pain specialist had always been \\known for his compassionate approach, but as opioid addiction rates climbed in the community {[}....{]}\end{tabular} \\ \hline\hline

\begin{tabular}[c]{@{}l@{}}Poor Sentence\\ Structure\end{tabular} & \begin{tabular}[c]{@{}l@{}}\textbf{\color{red}\st{As the night wore on, Z.'s laughter grew louder, his words slurring together like a sloppy melody. N. }}\\\textbf{\color{red}\st{and I exchanged a knowing glance, our concern simmering beneath the surface.}}\textbf{\color{ForestGreen}Z. was drinking more }\\\textbf{\color{ForestGreen}and more as the night went on. He laughed more loudly. His words started to slur, blurring one into} \\ \textbf{\color{ForestGreen} the next.I looked at N., who knew what I was thinking. We were going to have to take care of him.} At \\first, it was just  a slight stumble, a misstep that could be brushed off as a joke. But as the hours passed,{[}...{]}\end{tabular}                                                             \\ \hline\hline                                                                                                                                            \begin{tabular}[c]{@{}l@{}}Purple Prose\end{tabular} & \begin{tabular}[c]{@{}l@{}}My mother cried not just because twenty grand vanished into the ether{[}.....{]}.All of it vanished\textbf{\color{red}\st{, cycling back}} \\\textbf{\color{red}\st{through her mind, not as numbers but memories of scraped knees she bandaged alone and birthdays}}\\\textbf{\color{red}\st{ where her absence was felt more acutely than her presence.The sobs emerged from this deep well of}}\\\textbf{\color{red}\st{unspoken expectations, leaving behind a residue of weary resilience and a few hopeful echoes yet}}\\ \textbf{\color{red}\st{unwilling to completely extinguish.}} \textbf{\color{ForestGreen}She cried. She }\textbf{\color{ForestGreen} cried deep from this well of scraped knees she}\\\textbf{\color{ForestGreen} bandaged alone and birthdays she missed to work. She cried for unfairness. She cried without relief.}\end{tabular}
\\ \hline\hline
\begin{tabular}[c]{@{}l@{}}Awkward Word\\ Choice and \\Phrasing \end{tabular}  & \begin{tabular}[c]{@{}l@{}}I remember the city as a place of perpetual twilight, where the sky \textbf{\color{red}\st{seemed to hover}} \textbf{\color{ForestGreen}hovered} between dawn\\and dusk {[}....{]} glass towers, and the \textbf{\color{red}\st{sound}} \textbf{\color{ForestGreen}music} of sirens {[}.....{]}  bodega on the opposite side still \textbf{\color{red}\st{sold}} \textbf{\color{ForestGreen} reeked}\\\textbf{\color{ForestGreen}of} warm beer and stale cigarettes. The people were a blur of faces, each with their own story of [....] \end{tabular} \\ \hline\hline

\begin{tabular}[c]{@{}l@{}} Tense \\Inconsistency \end{tabular}&   \begin{tabular}[c]{@{}l@{}}As the sun dipped below the horizon, Elliot found himself engulfed by the growing darkness on Route 7. \\ The first snowflakes \textbf{\color{red}\st{began to drift}} \textbf{\color{ForestGreen}drifted} down from the heavens,{[}...{]}\end{tabular}      \\ \hline

\end{tabular}
\vspace{2ex}
\caption{\label{example_edits}Example of Edit types from our data}
\vspace{-5ex}
\end{table*}
With the finalized taxonomy of edits, we next conducted a larger-scale annotation study. The purpose of this study was to collect edits from writers on LLM-generated responses, categorizing them according to the established taxonomy. We validated the taxonomy's comprehensiveness by consulting writers who were not part of the formative study but participated in the editing task. These writers had the option to select an "Other" category and provide its name if an edit didn't fall into any of the seven established categories. The writers rarely chose the \textit{Other} category, doing so in only 10 out of 8,035 cases. These rare exceptions fell into categories such as ``Repetitive Sentence Structure," ``Confusing, Unclear or Incomplete Action/Meaning," and ``Mixed Metaphors". We also separately asked the writers if the categories encompassed all traits they encountered while editing these paragraphs and whether they would suggest any additions. Through email exchanges, all writers confirmed the taxonomy's comprehensiveness. 

This task followed a similar format to the formative study where participants were provided access to an editing interface (Figures \ref{interface}) populated with instructions and LLM-generated responses. In this interface, participants could select any span of text in the response and suggest a rewrite. Unlike the formative study, participants had to choose from the seven predefined categories in our taxonomy for each edit, rather than entering free-text categories. Participants received training about the taxonomy via email before beginning annotation. The training incorporated example edits for each category, akin to those in Table~\ref{example_edits}. Participants had no set limit on edits per response but were urged to improve the text as they saw fit. The interface logged all edits chronologically and offered an undo feature, enabling us to track the entire editing process, not just the final product.

After completing their edits, participants assigned two scores to the sample: an \textit{Initial Writing Quality Score} (IWQS) for the original response quality, and a \textit{Final Writing Quality Score} (FWQS) for the post-edit quality. Both used a 1-10 scale, with 1 being the lowest and 10 being the highest quality. The scores were incorporated to add a quantitative dimension to the qualitative process of editing. Additionally, the self-reported writing quality scoring system serves as a signal for writers to recognize their own improvements, set personal goals, and develop intrinsic motivation for enhancing their work.

Editing is a personal, time-consuming task, with edit quality dependent on participants having sufficient time to carefully read and consider improvements. To ensure quality, we maintained communication with all recruited participants. Participants completed the task in batches of 25, which typically took 3 hours, and were compensated \$100 USD for each batch. We recruited 18 writers with formal creative writing backgrounds from MFA mailing lists for our study, including 3 participants from the formative study (Table \ref{edit_participants}). Over 2.5 months, these writers edited LLM-generated responses based on their availability. Due to staggered start times, the number of edited samples varied among participants (see details in Table~\ref{edit_participants}). In total, each of the 1,057 <instruction, response> pairs we had prepared was edited by at least one participant, and 50 responses were edited by three participants, allowing us to study similarities and differences that occur when multiple writers edit the same response. The next section details the analysis we performed on the 8,000+ collected edits.

\section{The \corpus{}} \label{data}
\subsection{Overall Statistics}

We created the \corpus{} by collaborating with 18 writers who edited 1,057 LLM-generated paragraphs, gathering about 8 edits per paragraph, totaling 8,035 fine-grained edits. The data includes paragraphs from Claude3.5 Sonnet (368), GPT4o (393), and Llama3.1-70B (296). Figures~\ref{fig:data_analysis1}-\ref{fig:data_analysis2} present analyses of the \corpus{}, offering insights into how professional writers edit LLM-generated text and revealing a surprising lack of difference in writing quality across different model families \cite{zhou2024shared}.

\begin{table*}[!ht]
\centering
\small
\renewcommand{\arraystretch}{1.05}
\begin{tabular}{|l|l|l|}
\hline
\multirow{4}{*}{\begin{tabular}[c]{@{}l@{}}Meaning\\ Preserving\end{tabular}} & During the quarantine, the days stretched like endless corridors, each more indistinguishable from the last                                                            & \multirow{2}{*}{0.72} \\ \cline{2-2}
                                                                              & The days blurred into themselves during the quarantine, and I couldn't tell one from the other                                                                         &                       \\ \cline{2-3} 
                                                                              & She glanced down the hallway, suddenly aware of how quiet it was for a Tuesday evening                                                                                 & \multirow{2}{*}{0.79} \\ \cline{2-2}
                                                                              & It was eerily quiet for a Tuesday evening                                                                                                                              &                       \\ \hline
\multirow{4}{*}{\begin{tabular}[c]{@{}l@{}}Meaning\\ Changing\end{tabular}}   & \begin{tabular}[c]{@{}l@{}}Sophia took her smoking breaks in the back garden, a ritual she kept as precise as the time on\\  the old clock in her kitchen\end{tabular} & \multirow{2}{*}{0.52} \\ \cline{2-2}
                                                                              & One of the great comforts of old age was the ability to stop caring what other people thought                                                                          &                       \\ \cline{2-3} 
                                                                              & a brief reprieve from the unsaid words that floated between them                                                                                                       & \multirow{2}{*}{0.48} \\ \cline{2-2}
                                                                              & time to think of what to say. She hadn't told them her boyfriend was black                                                                                             &                       \\ \hline
\end{tabular}
\vspace{2ex}
\caption{\label{meaning}Examples of Meaning Preserving vs Meaning Changing Edits. Each example is a pair of original and edited span. Last Column shows the semantic similarity (BERT scores)}
\vspace{-5ex}
\end{table*}

\begin{table*}[!ht]
\small
\centering
\renewcommand{\arraystretch}{1.1}
\begin{tabular}{|l|c|c|c|}
\hline
         & W3                                                                                                                                            & W12                                                                                                                                                                                                                                                                & W16                                                                                                                                   \\ \hline
Original & \begin{tabular}[c]{@{}l@{}},the numbers glaring back at \\ me like an unsolvable riddle\end{tabular}                                          & \begin{tabular}[c]{@{}l@{}}glaring back at me like an unsolvable riddle\end{tabular}                                                                                                                                                                             & \begin{tabular}[c]{@{}l@{}},the numbers glaring back at \\ me like an unsolvable riddle\end{tabular}                                  \\ \hline
Category & Cliche                                                                                                                                        & Cliche                                                                                                                                                                                                                                                             & \begin{tabular}[c]{@{}l@{}}Unnecessary/Redundant \\ Exposition\end{tabular}                                                           \\ \hline
Edited   & . The numbers stared back                                                                                                                     & \begin{tabular}[c]{@{}l@{}}barreling over one another as they raced \\to\ some unseemly height.\end{tabular}                                                                                                                                                       & -                                                                                                                                     \\ \hline\hline
Original & \begin{tabular}[c]{@{}l@{}}and an unsettling sense of \\ mystery that gnawed at me \\ more than the inexplicable\\ weight itself\end{tabular} & \begin{tabular}[c]{@{}l@{}}Her words felt like a placeholder for an \\answer neither of us had yet. I walked out \\with a slip for blood tests and an unsettling\\ sense of mystery that gnawed at me more\\ than the inexplicable weight itself.\end{tabular} & \begin{tabular}[c]{@{}l@{}}-\end{tabular} \\ \hline
Category & Purple Prose                                                                                                                                  & Cliche                                                                                                                                                                                                                                                             & \begin{tabular}[c]{@{}l@{}}-\end{tabular}                                                           \\ \hline
Edited   & -                                                                                                                                             & \begin{tabular}[c]{@{}l@{}}But when I saw her turn to go, whispering\\ in the halls with a colleague, I knew there \\was still something she had yet to tell me.\end{tabular}                                                                                 & -                                                                                                                                     \\ \hline
\end{tabular}
\vspace{2ex}
\caption{\label{diffedits} Original spans selected by 3 writers from the same paragraph. `-' denotes the span was deleted while editing}
\vspace{-3ex}
\end{table*}

We analyze the editing process by examining \textit{edit operations}: insertion, deletion, or replacement. An edit is an insertion if it deletes no characters or adds 40+ characters net. Conversely, it's a deletion if it adds no characters or removes 40+ characters net. All other edits are tagged as replacements. We choose a threshold of 40 characters (roughly 10 words), to avoid labeling edits with minor length changes (``a'' to ``the'') as insertions or deletions. Figure~\ref{fig:data_plot1} shows edit operations by participant for each paragraph. Replacements are most frequent (74\%), followed by deletions (18\%) and insertions (8\%). Editing styles vary: some participants primarily use replacements (W2, W9, W10, W16, W18), while others employ deletions more often (W1, W5, W7, W8, W13, W17). Insertions are uncommon across all participants. To quantify meaning-preserving vs. meaning-changing edits, we calculate semantic similarity between original and edited text using BERT score\cite{zhang2019bertscore}. Using a threshold of 0.6 \footnote{This threshold was decided by manually analyzing 100 edits}, we classify edits with similarity > 0.6 as meaning-preserving. Of 6468 non-deletion edits, 70\% are meaning-preserving, with the rest meaning-changing. This finding supports our Design Principle 2. Figure \ref{fig:data_plot7} shows the distribution of the semantic similarity scores for the edits in LAMP. \footnote{However we note that with a threshold of 0.6, this appears to be quite a fuzzy/arbitrary distinction rather than a clear separation}. While the distribution is bimodal (with peaks around 0.6 and 0.75), these modes don't align well with the 0.6 threshold. This suggests that meaning preservation versus meaning change exists more on a continuum rather than as two clearly distinct categories.

The annotation interface allowed participants to provide Initial and Final Writing Quality Scores (IWQS and FWQS) for each paragraph, ranging from 1 to 10. Figure~\ref{fig:data_plot2} shows the distribution of these scores for each participant, revealing significant variability (e.g., W1's median IWQS is 7, W18's is 2). Calibration of writing quality scores is a known challenge, and we follow prior work in normalizing the scores into \textit{z-Scores} by subtracting the mean, dividing by the standard deviation of the scores for each participant \cite{graham2013continuous,maddela2022lens}, and re-scaling them to the 1 to 10 range. Subsequent analyses use these normalized scores.

We compute an \textit{edit distance} between the original LLM-generated text and the final edited text, by calculating a character-level Levenshtein distance \cite{Levenshtein1966BinaryCC} between the two strings of texts. The edit distance measures the ``amount of editing work'' performed by a writer. Figure~\ref{fig:data_plot4} shows a negative correlation between edit distance and IWQS (Pearson's $r = -0.31$), indicating that higher perceived text quality (high IWQS) requires less editing, while lower IWQS necessitates more editing.

Figure~\ref{fig:data_plot3} shows the average IWQS for each LLM on creative non-fiction and fiction writing tasks. Writers were unaware of which model generated each text, and tasks were shuffled to avoid bias. This analysis estimates the writing quality of the three models in both domains. Comparing model scores, we find no significant difference in writing quality across the three models. GPT-4o and Claude 3.5-Sonnet perform slightly better on creative non-fiction instructions (average 5.2) compared to Llama3.1-70B (5.0), though the difference is not statistically significant. All models show a slight decrease in performance for fictional instructions, with an average IWQS of 4.5. This suggests fiction writing may be more challenging for LLMs than creative non-fiction. These findings differ from task-oriented benchmarks that reveal performance gaps between models in areas like factual or logical reasoning. \textbf{Our results indicate that, when it comes to creative writing, writers perceived no significant qualitative differences among the texts generated by large language models (LLMs) such as GPT-4, Claude 3.5 Sonnet, and Llama 3.1 70B.}

Figure~\ref{fig:data_plot5} displays edit categories applied by writers to texts from three LLMs. The distribution is similar across models, with the most common categories being \textit{Awkward Word Choice and Phrasing} (28\%), \textit{Poor Sentence Structure} (20\%), \textit{Unnecessary/Redundant Exposition} (18\%), and \textit{Clichés} (17\%). Minor differences include GPT-4o using more purple prose and Llama3.1-70B generating more unnecessary exposition.
Overall, \textbf{LLMs across the three model families exhibit similar idiosyncrasies that are edited out in similar proportions by professional writers.} Figure~\ref{fig:data_plot6} illustrates the relationship between edit categories and IWQS. Higher IWQS scores correspond to fewer total edits, with texts rated 2 averaging 10.2 edits and those rated 10 receiving 2.4 edits, confirming that higher-quality texts need less editing. This trend however varies across edit categories: ``Unnecessary/Redundant Exposition'' and ``Lack of Specificity and Detail'' remain relatively constant, while the number of ``Awkward Word Choice and Phrasing'' and ``Cliché'' edits decrease as IWQS increases, suggesting a stronger correlation with perceived writing quality.

\subsection{\textit{Writers differ greatly in the amount of editing they do}: But to what extent?} \label{sec:writer_differences}

\begin{table*}[!ht]
\centering
\small
\renewcommand{\arraystretch}{1.05}
\begin{tabular}{|l|l|l|}
\hline
Syntactic Pattern & \begin{tabular}[c]{@{}l@{}}\% of \\Times\\      Edited\end{tabular} & Representative Sequence \\ \hline
DT NN IN NN CC    & 54\%                                                                   & \begin{tabular}[c]{@{}l@{}}a mix of pride and, a mix of fear and, a sense of protection and, a sense of wonder and, \\ a means of connection and, a pang of nostalgia and, a pang of disappointment, but, \\ a flicker of hope or, a blend of relaxation and,  a blend of curiosity and, the power of \\storytelling, the power of empathy, and, a web of belonging and {[}....{]}\end{tabular}                                                        \\ \hline
NN IN NN CC NN    & 35                                                                     & \begin{tabular}[c]{@{}l@{}}scene of chaos and destruction, mix of desperation and resolve, mix of relief and gratitude,\\ torn between curiosity and caution, breakfast of bread and jam, sense of calm and normalcy,\\ perception of loyalty and identity, glimmer of fear and vulnerability, meaning of protection \\and care, story of struggle and resilience, blend of fear and hope, {[}....{]}\end{tabular} \\ \hline
DT JJ NN IN PRP\$ & 40                                                                     & \begin{tabular}[c]{@{}l@{}}a constant reminder of his, the mundane routine of our, the intricate tapestry of its,\\ the subtle shift in their, the potential weight of its, a quiet sigh as her, \\ a small acknowledgment of their, the upcoming chapter of her, a silent battle between his, \\ a complex blend of their, the subtle shift in her, the unspoken plea in her {[}...{]}\end{tabular}                                                    \\ \hline
DT NN IN JJ NN    & 27                                                                     & \begin{tabular}[c]{@{}l@{}}the fabric of daily life, a moment of genuine connection, a life of absolute relaxation,\\ the face of inevitable loss, the weight of past grievances, a state of constant unease,\\ a residue of weary resilience, a sea of unspoken expectations, a mask of controlled concern,\\ the weight of unresolved history, a foundation of silent understanding, {[}....{]}\end{tabular}                                        \\ \hline
IN DT NN IN NN CC & 45                                                                     & \begin{tabular}[c]{@{}l@{}}with a mix of wariness and,  by the hum of traffic and,  in a flurry of pursuit and, \\ into a world of precision and, in a gesture of comfort and, in a storm of pain and,\\ for the sake of stability and, in the rhythm of routine or, in the magic of family and, \\ like the depth of understanding and, with a sense of nuance and {[}...{]}\end{tabular}                                                                 \\ \hline
\end{tabular}
\vspace{2ex}
\caption{\label{sequence}Idiosyncratic Sequences following certain syntactic patterns in LLM generated responses that are edited by writers. These syntactic patterns do not occur in the human written seed paragraphs}
\vspace{-7ex}
\end{table*}

\begin{figure*}[!ht]
\small
\centering
\includegraphics[width=0.75\textwidth]{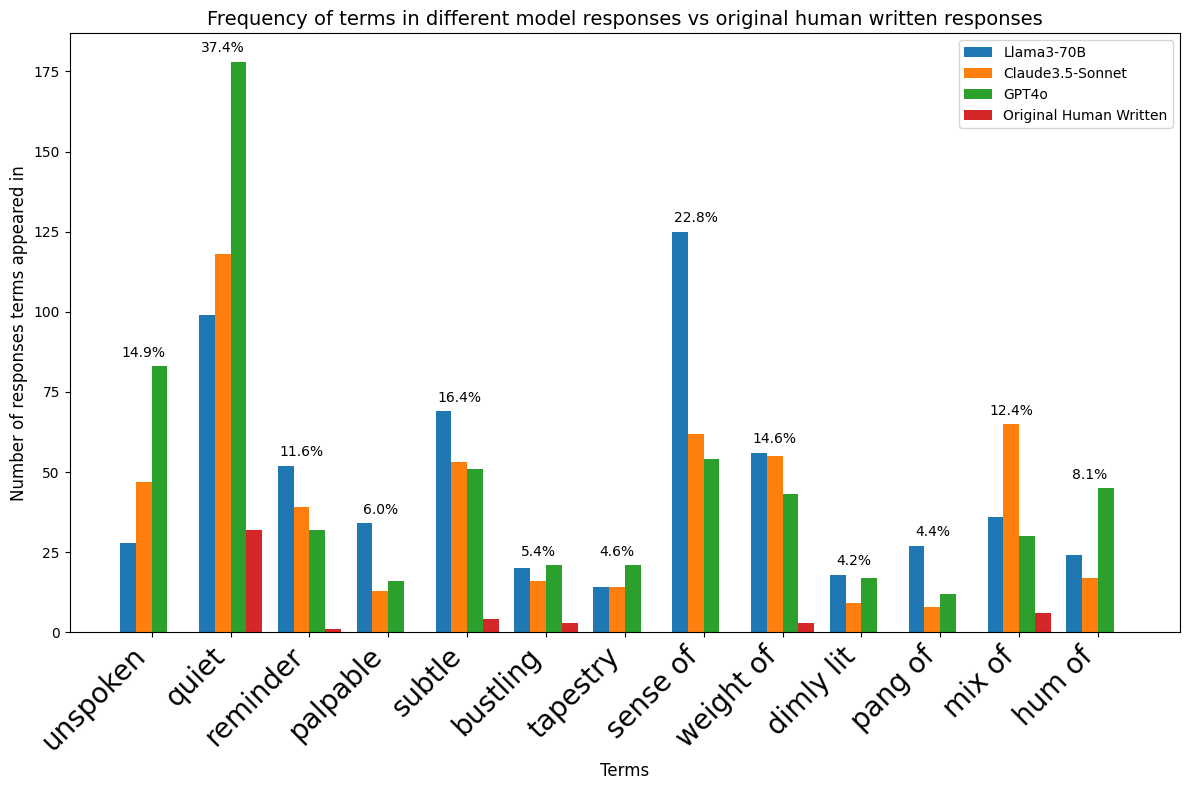}
\caption{\label{idiosyn}Distribution of peculiar and odd words and phrases occurring in LLM-generated text vs. human-written text in the \corpus{}.}
\end{figure*}

The writer's approaches to editing vary based on personal or organizational philosophy. Some prioritize preserving the original voice and make minimal changes to preserve authenticity \cite{tanselle1976editorial}. Others may take a more interventionist stance, heavily revising to align with their vision or house style. Additionally, some writers might make fewer but more impactful changes, while some might make numerous small revisions. To quantify this, we asked 3 writers (W3, W12, and W16) to edit a subset of the same 50 paragraphs from the \corpus{}. As expected, these three writers differed in the amount of editing they did. W3 did 9.4 edits on average while W12 and W16 did 6.0 and 6.3 edits on average. On average the span level precision (see Section \ref{detection} for more detail on the metric) between the 3 writers was 0.57 suggesting a moderately significant agreement.

Table \ref{diffedits} shows how sometimes writers select the exact same problematic span but assign different categories. A span that both W3 and W16 selected ``\textit{, the numbers glaring back at me like an unsolvable riddle}'' (Table \ref{diffedits} Row 1) was categorized differently (Table \ref{diffedits} Row 2). Yet both categorizations can be correct interpretations. When one relies on overused phrases or clichés, they often state the obvious or provide information that readers can easily infer implicitly. This results in redundant or superfluous exposition that doesn't add value to the narrative. Other times writers may select the same category but with only partial overlap on the selected span (Table \ref{diffedits} Row 1 W3 vs W12). Looking at (Table \ref{diffedits} Row 4 and 5; W3 vs W12) there is a partial overlap in the selected span ``\textit{and an unsettling sense of mystery that gnawed at me more than the inexplicable weight itself}". However, the selected categories are Purple Prose and Cliché respectively. Here again, it should be noted that Purple Prose is a style of writing that can be original or cliché, depending on its usage, context, and frequency. Not all elaborate writing is overused, but when certain ornate phrases or styles become too common, they can cross the line into cliché territory. W16 however did not edit this span.

We also highlight that diversity in edits among writers such as selecting different spans or rewriting it in an individualistic style is a positive aspect that prevents homogenization while still improving LLM-generated text as shown by our results in Section~\ref{rewriting}.

\subsection{Are there any specific stylistic idiosyncrasies in LLM generated responses?} \label{style}

Recent work from \citet{shaib2024detection} uses syntactic patterns with Part-of-speech \footnote{\url{https://www.sketchengine.eu/blog/pos-tags/}} as abstract representations of texts, that can capture more subtle repetitions than mere text memorization. They find that language models tend to use repetitive syntactic templates more often than humans and these patterns can help evaluate style memorization in language models. Following their experiments we consider Part-of-speech templates of length $n \epsilon \{5, 6, 7, 8\}$ in LLM-generated responses as well as the original seed human-written paragraphs (Table \ref{table:source_paragraph}). We looked at the 50 most common templates in LLM-generated responses and found that 15 templates do not occur as frequently in original human-written seed paragraphs. Table \ref{sequence} shows representative sequences corresponding to particular syntactic patterns present in higher proportion in LLM-generated responses. These sequences constitute categories of \textbf{Clichés, Unnecessary/Redundant Exposition or Poor Sentence Structure} and are often heavily edited by writers in our study.

To better understand idiosyncrasies, we examined awkward words/phrases occurring disproportionately in LLM-generated responses. For instance, Figure \ref{idiosyn} shows how a word like \textit{unspoken} occurs in about 15\% of LLM-generated responses. Similarly phrases such as \textit{weight of}, \textit{sense of}, \textit{mix of} occur very rarely or not at all in original seed paragraphs (Table \ref{table:source_paragraph}) while they occur frequently in LLM-generated responses. We also found peculiar and uncommon phrases generated by LLMs across several responses such as \textit{air was thick}, \textit{hung in the air}, \textit{eyes darting}, \textit{a sense of unease (grew/growing/settles) in the pit of (her/my) stomach}. The most surprising finding is that all 3 LLMs generate these idiosyncratic words/phrases \textbf{suggesting possible overlap/mixture in instruction tuning data across model families or one model trained on synthetic data generated from another model} \cite{zhou2024shared}. 

\section{AUTOMATIC DETECTION AND REWRITING OF LLM IDIOSYNCRASIES} \label{automatic}

While several automated editing approaches exist for improving LLM outputs at scale, we evaluate their effectiveness specifically for addressing idiosyncrasies in creative writing. Building on \citet{hayes1987cognitive} and \citet{scardamalia1983development}, we develop techniques to separate detection and rewriting tasks, evaluating them using \corpus{} annotations. Given automated evaluation limitations for text editing\cite{dou2023automatic}, we conduct a large-scale preference annotation study with \corpus{} writers, comparing human and LLM-produced edits. To accommodate methods that require training samples, we split our data: 146 of 1057 \corpus{} paragraphs for training, the rest for testing.

\subsection{Automatic detection of problematic spans in LLM-generated text} \label{detection}

We formulate the problem of detecting problematic spans in LLM-generated text as a \textit{multi-span categorical extraction} problem. In other words, given a paragraph of LLM-generated text, the method must output a list of non-overlapping spans present in the original text, and assign a category to each extracted span (from the list of categories of the \corpus{}).

To evaluate various methods, we use the span-level precision metric, a common metric used in NLP tasks requiring comparison of extracted spans \cite{rajpurkar2016squad}. Span-level precision measures the degree of overlap between predicted spans and reference or ground truth spans (in our case spans collected as a part of the editing process from writers). The overlap is measured at the character level, such that spans that partially overlap will get precision scores that reflect the amount of overlap between the two spans. High span-level precision indicates that the model is precise in identifying the correct boundaries of relevant text spans without over-predicting. We implement two precision metric variants: General and Categorical Precision. General Precision credits span selection regardless of category assignment, while Categorical Precision requires correct category assignment. We use a precision-based metric (like BLEU \cite{papineni2002bleu}) rather than recall-based (like ROUGE \cite{lin2004rouge}) as LLM-based methods tend to over-generate spans, which recall doesn't penalize. Our focus is measuring the overlap between generated and ground truth spans. Appendix~\ref{sec:precision_explanation} provides a simplified example where General and Categorical Precision are computed, illustrating the suitability of these metrics for this evaluation setting.

\begin{table}[!ht]
    \begin{tabular}{lrrrrrr}
        % \toprule
        & \multicolumn{3}{c}{\textbf{General}} & \multicolumn{3}{c}{\textbf{Categorical }} \\
         \cmidrule(l){2-4} \cmidrule(l){5-7}
        Expert Agreement & \multicolumn{3}{c}{0.57} & \multicolumn{3}{c}{0.23} \\
        \cmidrule(l){1-1} \cmidrule(l){2-4} \cmidrule(l){5-7}
         \textbf{Detector LLM} & n=2 & n=5 & n=25 & n=2 & n=5 & n=25 \\
        \midrule
         Claude3.5-Sonnet & 0.43 & \textbf{0.46} & 0.44 & 0.20 & 0.21 & 0.20 \\
         Llama3.1-70b & 0.42 & 0.45 & 0.38 & 0.16 & 0.21 & 0.14 \\
         GPT-4o & 0.44 & \textbf{0.46} & \textbf{0.46} & 0.17 & 0.18 & 0.17 \\
        \bottomrule
    \end{tabular}
    \vspace{2ex}
    \caption{Results of automated methods for detecting problematic spans in LLM-generated text, compared to agreement levels between three experts. Results report Precision scores for various LLMs used for detection, using a {2,5,25} examples in the few-shot prompt instruction, reporting both a General Precision and the stricter Categorical Precision.}
    \label{tab:detection_results}
\vspace{-5ex}
\end{table}

\begin{table*}[!ht]
\centering
\small
\renewcommand{\arraystretch}{1.15}
\begin{tabular}{|l|l|}
\hline
Writer & \begin{tabular}[c]{@{}l@{}}Jackson leaned back in his office chair [$\textbf{\color{red}, staring out}_{\textsc{\textbf{\color{blue}Poor Sentence Structure}}}$] through the expansive glass windows\\of the high-rise building.
Below him, the city churned with [$\textbf{\color{red} relentless}_{\textsc{\textbf{\color{blue}Awkward Word Choice and Phrasing}}}$] energy, \\ $[\textbf{\color{red} a tapestry of modernity threaded with the hum of traffic,}_{\textsc{\textbf{\color{blue}Cliché}}}]$ the glow of digital billboards, $[$\textbf{\color{red}and the}\\ $\textbf{\color{red} unceasing flow of pedestrians. His desk, a}_{\textsc{\textbf{\color{blue}Poor Sentence Structure}}}]$ chaotic mixture of reports and coffee cups \\
$[\textbf{\color{red}, contrasted sharply with the sleek orderliness of the cityscape}_{\textsc{\textbf{\color{blue}Poor Sentence Structure}}}]$ [ \textbf{\color{red}Despite the} \\ $\textbf{\color{red}apparent advancement that framed his daily life, Jackson felt an inexplicable disconnection.}$\\  $\textbf{\color{red}His}_{\textsc{\textbf{\color{blue}Unnecessary/Redundant Exposition}}}]$ mind often wandered to an untouched box of old photographs at home, \\depicting simpler times – wooden houses, dirt paths, faces etched with stories of a bygone era. Those images \\spoke to him more than the polished chrome and glass now surrounding him.[ \textbf{\color{red}He couldn't help but yearn for the}\\ \textbf{\color{red}intangible warmth of community gatherings under open skies, unhurried conversations, and the tangible}\\\textbf{\color{red} reality of a slower pace. Here, in the heart of progress, he was a mere observer, detached from the legacy}\\ $\textbf{\color{red}beneath the steel and concrete, his own inclinations buried beneath the layers of modernity.}_{\textsc{\textbf{\color{blue}Cliché}}}]$\end{tabular}\\ \hline
LLM    & \begin{tabular}[c]{@{}l@{}}
Jackson leaned back in his office chair, staring out through the expansive glass windows of the high-rise building.\\ Below him, the city [$\textbf{\color{red} churned with relentless energy}_{\textsc{\textbf{\color{blue}Cliche}}}$].The city was [ \textbf{\color{red} a tapestry of modernity threaded} \\ $\textbf{\color{red} with the hum of traffic, the glow of digital billboards, and the unceasing flow of pedestrians}_{\textsc{\textbf{\color{blue}Purple Prose}}}$]. \\His desk, a [$\textbf{\color{red} chaotic mixture of reports and coffee cups}_{\textsc{\textbf{\color{blue}Cliche}}}$] contrasted sharply with \textbf{\color{red}the sleek orderliness.}\\$[\textbf{\color{red} of the cityscape}_{\textsc{\textbf{\color{blue}Cliche}}}]$. Despite the apparent advancement that framed his daily life, Jackson felt \textbf{\color{red}an inexplicable} \\$[\textbf{\color{red} disconnection}_{\textsc{\textbf{\color{blue}Cliche}}}]$ His mind often wandered to an untouched box of old photographs at home, depicting simpler\\ times – $[\textbf{\color{red} wooden houses, dirt paths, faces etched with stories of a bygone era.}_{\textsc{\textbf{\color{blue}Cliche}}}]$ He couldn't help but yearn\\ for $[\textbf{\color{red}the intangible warmth of community gatherings under open skies, unhurried conversations, and the}$ \\ $\textbf{\color{red}tangible reality of a slower pace}_{\textsc{\textbf{\color{blue}Purple Prose)}}}]$ Here, in the $[\textbf{\color{red}heart of progress}_{\textsc{\textbf{\color{blue}Cliche}}}]$, he was a mere observer, \\ $[$\textbf{\color{red}detached from the legacy beneath the steel and concrete, his own inclinations buried beneath the layers of}\\$\textbf{\color{red} modernity.}_{\textsc{\textbf{\color{blue}Purple Prose)}}}]$
\end{tabular}
 \\ \hline
\end{tabular}
\vspace{2ex}
\caption{\label{detection_contrast}Example of problematic spans identified by a writer and by a LLM.}
\end{table*}

We implement few-shot LLM-based methods \cite{NEURIPS2020_1457c0d6} that have demonstrated competitive performance on tasks across several disciplines, often using fewer than 100 examples \footnote{Model fine-tuning experiments on \corpus{} are left for future work}. Our experiment varies the number of few-shot examples (2, 5, and 25) with the 2-shot prompt in Appendix~\ref{app:detection_prompt} and tests Llama3.1-70B, GPT-4o, and Claude3.5-Sonnet. As part of the collection of the \corpus{}, 50 paragraphs were edited independently, and we computed General and Categorical Precision on this set to estimate expert-expert agreement. Table~\ref{tab:detection_results} summarizes results. The best General Precision (0.46) is achieved by Claude-3.5 Sonnet and GPT-4o with a 5-shot prompt, below the expert agreement level (0.57). LLM-based methods can identify problematic spans with significant expert overlap, but improvement is possible. Performance improves from 2-shot to 5-shot prompts but plateaus thereafter (Claude-3.5 Sonnet and GPT-4o achieve similar or better performance with 5-shot vs. 25-shot prompts).

Categorical Precision is consistently lower than General Precision for both LLM-based methods and writer-writer comparison, suggesting that even when problematic spans are commonly identified, category agreement may differ. Section~\ref{sec:writer_differences} explores this finding in detail, with examples of spans that could be reasonably assigned multiple categories. Table \ref{detection_contrast} illustrates the contrast between automatic editing and writer-selected edits by showing a paragraph with problematic spans and categories identified by a writer versus an LLM.

\subsection{Automatic rewriting of problematic spans in any LLM-generated text} \label{rewriting}

\begin{table*}[!ht]
\centering
\small
\renewcommand{\arraystretch}{1.13}
\begin{tabular}{|l|l|}
\hline
\begin{tabular}[c]{@{}l@{}}Writer\\ Detected\\ and LLM \\Rewritten\end{tabular} & \begin{tabular}[c]{@{}l@{}}Dust settled on my window sill, \textbf{\color{red}\st{a quiet testament to the daily upheaval outside}} \textbf{\color{ForestGreen}a reminder of the} \\\textbf{\color{ForestGreen}ongoing transformation outside}. Yellow machines trundled along the narrow street, \textbf{\color{red}\st{their relentless}}\\ \textbf{\color{red}\st{growl a constant soundtrack}}
\textbf{\color{ForestGreen} growling incessantly}. I watched the workers in their neon vests, \textbf{\color{red}\st{like}}\\\textbf{\color{red}\st{bright insects against the concrete hive}} \textbf{\color{ForestGreen}bright and busy against the concrete backdrop}, orchestrating\\ a ballet of drills and hammers. Every so often, a stone \textbf{\color{red}\st{house crumbled}} \textbf{\color{ForestGreen}mansion fell} to give way to sleek \\modernism, steel skeletons reaching shyly towards a haze-obscured sky. \textbf{\color{red}\st{The trees lining the avenue,}}\\ \textbf{\color{red}\st{long sentries guarding our history, stood fewer each day}} \textbf{\color{ForestGreen}The trees lining the avenue dwindled} \\\textbf{\color{ForestGreen} each day}. Neighbors \textbf{\color{red}\st{nodded at one another with tight smiles}} \textbf{\color{ForestGreen}exchanged tight smiles,} masking the void \\\textbf{\color{red}\st{of vanished gardens and familiar creeks}} \textbf{\color{ForestGreen}left by vanished gardens}. \textbf{\color{red}\st{Change buzzed in the air, metallic}}\\ \textbf{\color{red}\st{and cold, seeping into conversations and dreams.}} \textbf{\color{ForestGreen}Change felt cold and metallic, entering}\\ \textbf{\color{ForestGreen}conversations and dreams}. I \textbf{\color{red}\st{found myself wondering if we were}} \textbf{\color{ForestGreen}wondered if we were} building towards\\ \textbf{\color{red}\st{something}}\textbf{\color{red}\st{ or away from it}}\textbf{\color{ForestGreen}something or abandoning it.}\end{tabular} \\ \hline
\begin{tabular}[c]{@{}l@{}}LLM\\ Detected\\ and LLM \\Rewritten\end{tabular}    & \begin{tabular}[c]{@{}l@{}}\textbf{\color{red}\st{Becky's mind races, her thoughts a tangled mess of worries and what-ifs.}} \textbf{\color{ForestGreen}Becky's thoughts spiral}.\\She knows she should eat, but \textbf{\color{red}\st{the mere thought of food turns her stomach.}}\textbf{\color{ForestGreen} she has no appetite}. Instead, \\ she \textbf{\color{red}\st{finds herself pacing the worn carpet of her small apartment, her fingers absently tracing the edges}}\\ \textbf{\color{red}\st{of picture frames and trinkets.}} \textbf{\color{ForestGreen}paces her small apartment, fingers tracing picture frames and trinkets} \\ \textbf{\color{ForestGreen}as she walks.} She pauses at the window, watching the world outside \textbf{\color{red}\st{continue its relentless march forward}}\\ \textbf{\color{ForestGreen}go about its day.} A neighbor walks their dog, \textbf{\color{red}\st{oblivious to her inner turmoil}} \textbf{\color{ForestGreen}oblivious}. Becky's gaze drifts\\ to the potted plants on her windowsill, their leaves drooping slightly. She reaches for the watering can,\\ \textbf{\color{red}\st{tending to their needs as a way to quiet her own}}\textbf{\color{ForestGreen}nurturing them to distract herself}.\textbf{\color{red}\st{The simple act of}}\\\textbf{\color{red}\st{ nurturing something else helps ground her, if only momentarily.}} \textbf{\color{ForestGreen}The simple act grounds her}\\ \textbf{\color{ForestGreen} momentarily.} She moves to her bookshelf, running her fingertips along the spines of well-loved novels, \\\textbf{\color{red}\st{searching for a familiar story to lose herself in}} \textbf{\color{ForestGreen}hoping to find comfort in a familiar story.} As she settles \\into her favorite chair, book in hand, she notices her heartbeat has slowed. \textbf{\color{red}\st{The knot in her stomach loosens,}}\\\textbf{\color{red}\st{ ever so slightly.}} \textbf{\color{ForestGreen}She exhales slowly.} She may not be ready to eat yet, but \textbf{\color{red}\st{she's found a moment of peace in}}\\ \textbf{\color{red}\st{the chaos of her mind.}} \textbf{\color{ForestGreen}she's found a small respite.}\end{tabular}
 \\ \hline
\end{tabular}
\vspace{2ex}
\caption{\label{AIedit} Table showing LLM-edited paragraphs on both Oracle Writer predicted spans as well as LLM predicted spans}
\vspace{-3ex}
\end{table*}

To propose improvements for detected problematic spans, we use few-shot prompting \cite{NEURIPS2020_1457c0d6} with LLMs as well. We design prompts for each of the seven edit categories, incorporating examples of rewrites from writers. Each prompt includes a category definition, 25 examples from the \corpus{} with original paragraphs, a single problematic span for the category, the expert-proposed rewrite, and finally the input paragraph with the target span to be rewritten. The prompts for each category are listed in Appendix \ref{rewriting_prompts}.

The detection and rewriting methods can form a two-step pipeline for editing paragraphs. Detection identifies problematic spans and assigns categories while rewriting uses category-specific prompts to revise each detected span. A final step replaces all problematic spans in the original paragraph with their rewrites. Unlike the detection task, we do not evaluate the rewriting stage in isolation. Instead, we judge the complete pipeline that edits an entire paragraph (by detecting and rewriting multiple spans) through manual evaluation with 12 writers that annotated the \corpus{}. We describe this manual experiment next.

\subsection{Evaluating Automatic Editing of LLM-generated Text} \label{auto:rewrite}

To evaluate editing quality, we design an evaluation task where participants read three variants of a paragraph and rank them in terms of overall preference: (1) an unedited \textbf{LLM-generated} paragraph from the \corpus{}, (2) the \textbf{Writer-edited} version from the \corpus{}, and (3) an \textbf{LLM-edited} version using our pipeline to detect and rewrite problematic spans. 6 out of 18 of our experts were unavailable during the preference annotations. Some of them were busy with their full-time jobs while others wanted to take a break after the edit task. Given the concern for proper turn around we re-hired 12 of the 18 experts who had participated in creating the \corpus{} for this evaluation.

We split the LLM-edited variant into two further sub-conditions:
\begin{itemize}
    \item \textbf{Writer Detected and LLM Rewritten}: In this condition, the pipeline skips automatic detection of problematic spans, relying only on reference spans selected by the writer during manual editing. It runs solely the rewriting stage, simulating an oracle setting where problematic spans are manually provided. This condition is coded as \textbf{LLM-edited-Oracle}
    \item \textbf{LLM Detected and LLM Rewritten}: In this condition, the two-step pipeline is entirely automatic, with the automatically detected spans being provided to the automatic rewriting module. This condition fully automates editing of the paragraph and is coded as \textbf{LLM-edited-full}.
\end{itemize}

\begin{figure*}[!ht]
    \centering
    \frame{
        \includegraphics[width=0.95\textwidth]{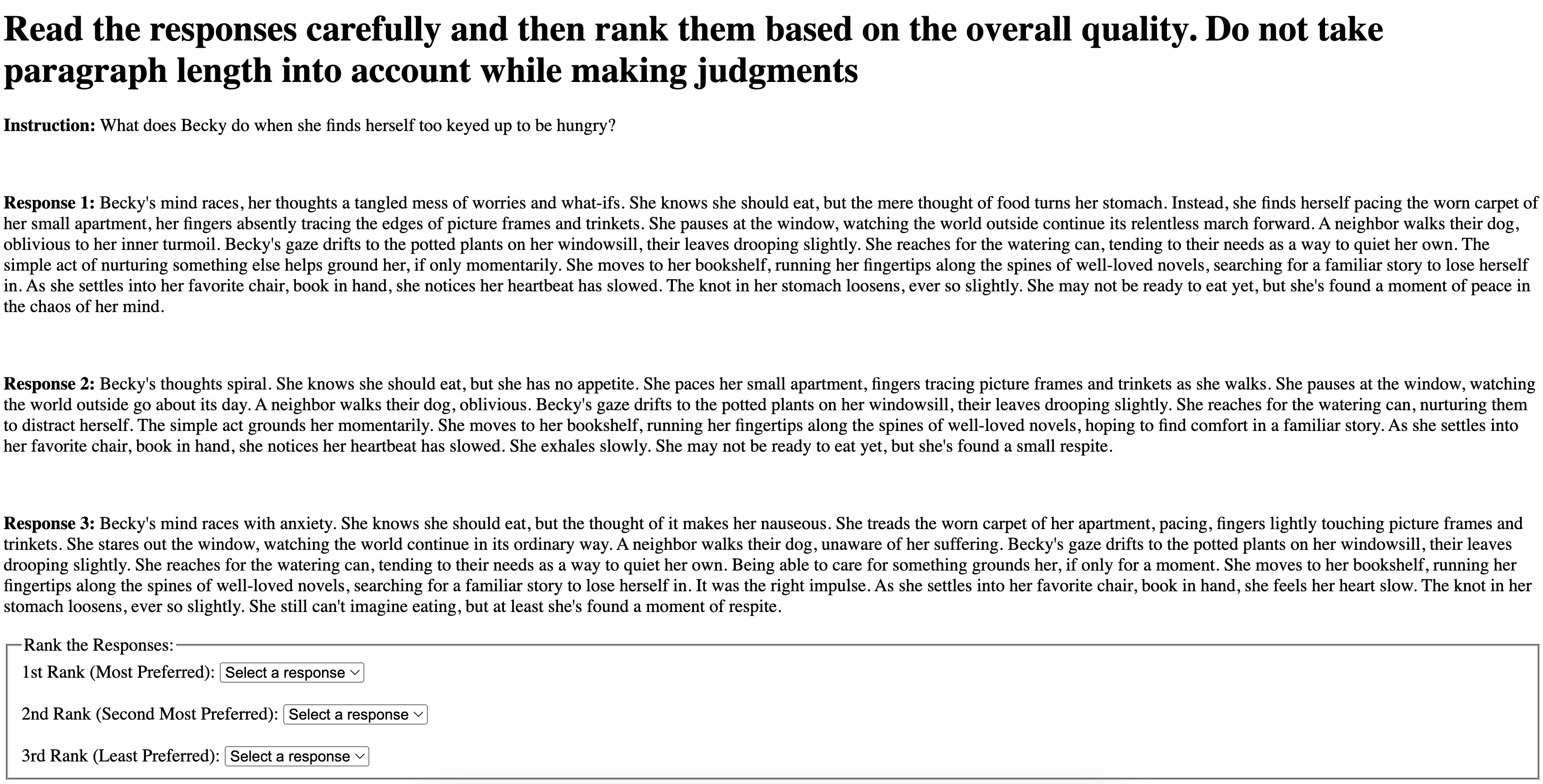}
    }
    \caption{\label{preference_interface} Interface used by participants to read through variants of a paragraph (one LLM-generated, one manually edited by an expert, one edited by an LLM-based system), and rank them in terms of preference.}
\end{figure*}

\begin{table*}[!ht]
    \small
    \centering
    \renewcommand{\arraystretch}{1.15}
    \begin{tabular}{lcc}
    \multicolumn{1}{l|}{LLM-generated} & \multicolumn{1}{c|}{Writer-edited} & LLM-edited-full \\ \hline
    \multicolumn{1}{l|}{2.55}         & \multicolumn{1}{c|}{1.47}      & \multicolumn{1}{c}{1.99}           \\ \hline\\
    \multicolumn{1}{l|}{LLM-generated} & \multicolumn{1}{c|}{Writer-edited} & LLM-edited-oracle \\ \hline
    \multicolumn{1}{l|}{2.47}         & \multicolumn{1}{c|}{1.53}      & \multicolumn{1}{c}{1.99}          \\ \hline
    \end{tabular}

\vspace{2ex}
\caption{\label{tab:preference_results}Average Ranking across 600 preference judgments. LLM-edited $>$ LLM-generated (p-value: 1.3e-11 for Writer Predicted spans;  2.8e-13 for LLM Predicted spans) and Writer-edited $>$ LLM-generated (p-value: 1.1e-26 for Writer Predicted spans;  1.17e-31 for LLM Predicted spans) using Wilcoxon signed-rank test}
\vspace{-3ex}
\end{table*}

\begin{figure*}
    \subfloat[Samples involving oracle detection (N=300)]{\includegraphics[width=0.47\textwidth]{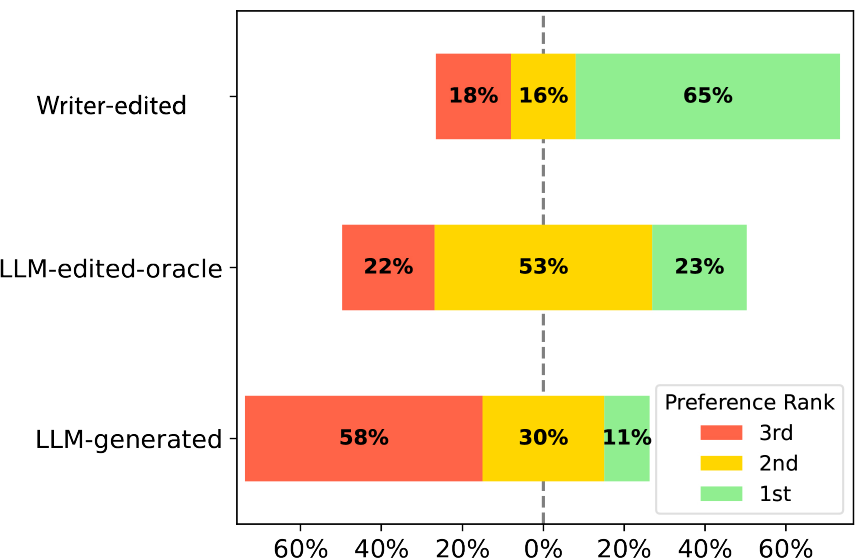}} \quad
    \subfloat[Samples involving automated detection (N=300)]{\includegraphics[width=0.47\textwidth]{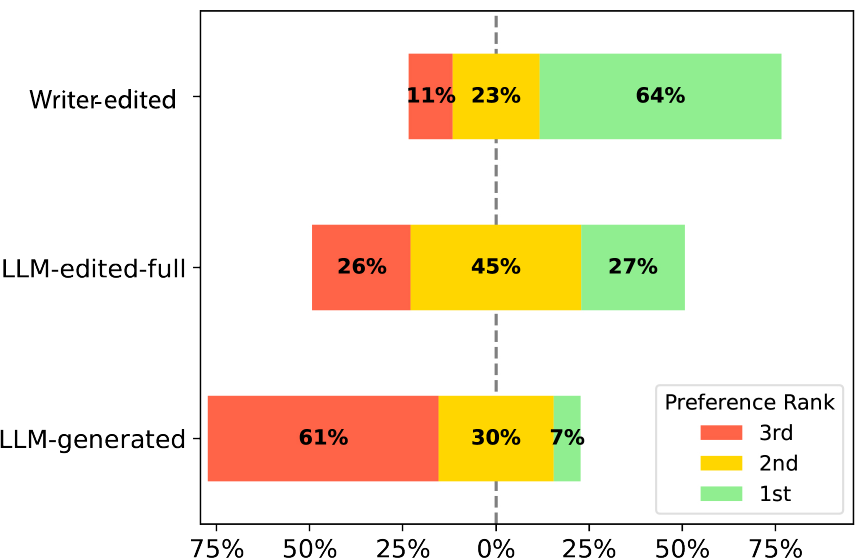}} 
    \caption{\label{preference_graphs} Distribution of rankings for each variant in the preference annotation study. Annotators read three variants of a paragraph (Writer-edited, LLM-generated, and either LLM-edited-oracle or LLM-edited-full) and ranked them by preference (1st, 2nd, 3rd). The distribution indicates how often each variant was ranked as best (1st), second best (2nd), or worst (3rd).}
\end{figure*}

Table \ref{AIedit} shows examples of LLM-edited paragraphs under both sub-conditions. Although the examples here showcase `Replacement' and `Deletion'  edits, analysis on a larger corpus of 200 LLM-edited paragraphs confirms that the automatic editing pipeline mirrors the edit type distribution of expert editors (shown in Figure~\ref{fig:data_plot1}): roughly 65\%  `replace` edits, 25\% `delete` edits, and 10\% `insert` edits. This surface-level analysis confirms the automatic pipeline's ability to mimic the editing process of experts measured in terms of relative proportions of character insertion and deletion.

In our pilot evaluation, we initially included all four conditions for annotation. However, ranking four paragraphs proved challenging for participants, especially when distinguishing between their second and third preferences. Based on this feedback, we redesigned the task to have participants judge only three conditions in each annotation. We always included the \textbf{LLM-generated} and \textbf{Writer-edited} paragraphs and alternated between including \textbf{LLM-edited-oracle} and \textbf{LLM-edited-full} paragraphs. While traditionally pairwise rankings tend to be easier for humans to annotate, three-way ranking allowed direct comparison between the key conditions we wanted to evaluate (LLM-generated v.s. Writer-edited v.s. LLM-edited), making the relative preferences clearer in a single evaluation rather than having to combine multiple pairwise comparisons. Additionally, we want to avoid any familiarity effects or biases that could arise from evaluators reading the same text multiple times, which would likely occur when collecting overlapping pairwise comparisons.

We note that to obtain automatic edits of a paragraph, we used the same LLM that had originally been used to generate the paragraph.\footnote{In other words, we used GPT-4o in the two-step pipeline to generate edits to paragraphs that were originally generated by GPT-4o.} While not optimal, as a single LLM might offer slightly better detection and rewriting capabilities, this approach allows us to simplify the experiment conceptually and also test our hypothesis if edits lead to overall better alignment without relying on a single model family. We assess if using an LLM in a multi-stage pipeline (drafting, problem detection, rewriting) can enhance overall writing quality. Future work could potentially optimize this editing pipeline further, possibly yielding better results for LLM-edited conditions.

To ensure fairness, paragraph variants are displayed in a shuffled order and anonymized, and participants were not informed about the difference between the paragraphs (i.e. whether they are edited). For the curious reader, Figure~\ref{preference_interface} provides the interface used for the annotation task, including three variants of a paragraph. To conduct our experiments, we selected a total of 200 paragraph triplets (100 including an \textbf{LLM-edited-oracle} paragraph, and 100 including an \textbf{LLM-edited-full} paragraph) selecting samples from the \corpus{}'s test set. Preference judgments were collected in batches of 25-35 paragraph triplets, with participants paid \$35/hour. To account for potential subjectivity and calculate agreement and reliability, three experts judged each triplet, totaling 600 annotated preference rankings. To ensure the validity of the results, \textbf{no participant reviewed paragraphs they had seen or edited in past tasks, and only judged paragraphs edited by other experts}. 

To analyze the reliability of the results we calculate inter-annotator agreement using Kendall's W (also known as Kendall's coefficient of concordance) \cite{field2005kendall} which ranges from 0 (no agreement) to 1 (complete agreement) to evaluate agreement amongst participants. Our annotation achieves an overall agreement of \textbf{0.505}, suggesting a moderate level of agreement across all participants. This moderate agreement underscores the subjective nature of judging writing quality while suggesting that certain differences are distinctive enough to be consistently preferred by multiple participants.

Table~\ref{tab:preference_results} and Figure~\ref{preference_graphs} summarize the preference evaluation results, showing average ranks across 600 annotations. Overall, the \textbf{Writer-edited condition is most preferred, a sign that expert-edited text is unrivaled in terms of writing quality}, being marked as the most-preferred paragraph variant 65\% of the time and achieving an average rank of 1.5. Next, the LLM-edited variants come in second, with an average rank of 1.99 for both the LLM-edited-oracle and LLM-edited-full conditions. Surprisingly, the condition that leveraged the oracle span from writers ranks almost identical to the condition with automatically detected spans. This provides evidence that detection of problematic spans is not the bottleneck in improving writing quality, and instead the \textbf{rewriting module (which is common to both conditions) is what dictates the overall performance of an automated text-editing pipeline}. Finally, the original \textbf{LLM-generated} paragraphs achieve the worst ranking performance, being least preferred 60\% of the time, and achieving an average rank of 2.51.

In summary, our experiment validates the potential benefit of automatic editing to improve writing quality: although automatic editing does match the quality of edits provided by professional writers, LLM-edited text is significantly preferred to LLM-generated text by expert writers (Design Principle 3). In other words, this experiment shows that \textbf{LLMs can improve the quality of their writing in a fully automatic way, by first generating a draft, selecting problematic spans, and then rewriting such spans}. These results align with previous findings \cite{wadhwa2024learning,pan2024automatically,gao2024aligning,dwivedi2022editeval} showing that iterative refinement and editing can improve LLM outputs.

\begin{table*}[!ht]
\renewcommand{\arraystretch}{1.05}
\centering
\small
\begin{tabular}{|l|l|l|l|}
\hline
Genre                                                                       & Category                                                                             & Original Span                                                                                                                                                                                                                                                     & LLM edited span                                                                                                                                                                                                                                              \\ \hline
\begin{tabular}[c]{@{}l@{}}Literary\\ Fiction\end{tabular}                  & \multirow{2}{*}{\begin{tabular}[c]{@{}l@{}}Lack of \\ Specificity\\ and Detail\end{tabular}} & \begin{tabular}[c]{@{}l@{}}seems to embody the city's frustrations\\ and disappointments\end{tabular}                                                                                                                                                             & \begin{tabular}[c]{@{}l@{}}seems to embody the city's frustrations \\ and disappointments, his eyes a deep \\ well of anger and desperation, his face a \\ topographic map of the city's corruption \\ and neglect.\end{tabular}                             \\ \cline{1-1} \cline{3-4} 
\begin{tabular}[c]{@{}l@{}}Food \\ Writing\end{tabular}                     &                                                                                              & \begin{tabular}[c]{@{}l@{}}Gone are the individual ramekins, \\ replaced by a single, generous vessel \\ that invites sharing.\end{tabular}                                                                                                                       & \begin{tabular}[c]{@{}l@{}}Gone are the individual ramekins, replaced \\ by a deep-dish pie plate that invites \\ communal indulgence. The golden crust \\ cradles a sea of silky custard, its surface\\ a glassy expanse of caramelized sugar.\end{tabular} \\ \hline\hline
\begin{tabular}[c]{@{}l@{}}Literary\\ Fiction\end{tabular}                  & \multirow{3}{*}{\begin{tabular}[c]{@{}l@{}}Purple \\ Prose\end{tabular}}                     & \begin{tabular}[c]{@{}l@{}}confusion clouded his understanding \\ like the haze hanging over the park\end{tabular}                                                                                                                                                & he remained confused                                                                                                                                                                                                                                         \\ \cline{1-1} \cline{3-4} 
\begin{tabular}[c]{@{}l@{}}Food\\ Writing\end{tabular}                      &                                                                                              & \begin{tabular}[c]{@{}l@{}}It's the democratization of ingredients \\ that sets the bánh mì apart\end{tabular}                                                                                                                                                    & \begin{tabular}[c]{@{}l@{}}It's the variety of ingredients that \\ sets the bánh mì apart\end{tabular}                                                                                                                                                       \\ \cline{1-1} \cline{3-4} 
\begin{tabular}[c]{@{}l@{}}Internet\\ Advice\end{tabular}                   &                                                                                              & \begin{tabular}[c]{@{}l@{}}But pain has a way of cracking us open,\\ making space for new possibilities we \\ couldn't have imagined before.\end{tabular}                                                                                                         & \begin{tabular}[c]{@{}l@{}}But pain can open us up to new \\ possibilities.\end{tabular}                                                                                                                                                                     \\ \hline\hline
\begin{tabular}[c]{@{}l@{}}Travel\\ Writing\end{tabular}                    & \multirow{2}{*}{\begin{tabular}[c]{@{}l@{}}Poor\\ Sentence\\ Structure\end{tabular}}         & \begin{tabular}[c]{@{}l@{}}He tips generously but modestly, as if \\ acknowledging a pact of respect\end{tabular}                                                                                                                                                 & \begin{tabular}[c]{@{}l@{}}He tips generously yet with restraint,\\ honoring a mutual respect\end{tabular}                                                                                                                                                   \\ \cline{1-1} \cline{3-4} 
\begin{tabular}[c]{@{}l@{}}Literary\\ Fiction\end{tabular}                  &                                                                                              & \begin{tabular}[c]{@{}l@{}}The gulf between her vision for her \\ future and her parents' hopes wasn't \\ something that could be bridged with \\ words alone, and Bella realized that \\ sometimes growth requires difficult\\  choices.\end{tabular}            & \begin{tabular}[c]{@{}l@{}}Bella realized that her vision for the \\ future and her parents' hopes were \\ irreconcilable. Growth, she understood,\\ often demands difficult choices.\end{tabular}                                                           \\ \hline\hline
\begin{tabular}[c]{@{}l@{}}Internet\\ Advice\end{tabular}                   & \multirow{2}{*}{\begin{tabular}[c]{@{}l@{}}Unnecessary\\ Exposition\end{tabular}}            & , a silent reminder of everything it once meant                                                                                                                                                                                                                   & -                                                                                                                                                                                                                                                            \\ \cline{1-1} \cline{3-4} 
\begin{tabular}[c]{@{}l@{}}Personal\\ Essay\end{tabular}                    &                                                                                              & \begin{tabular}[c]{@{}l@{}}Now, when life deals its inevitable blows, I \\ think of those stones. I remember that \\ sometimes it's the cracks that let the light in, \\ and that even the most jagged edges can be\\ smoothed by time and patience.\end{tabular} & \begin{tabular}[c]{@{}l@{}}Now, when life deals its inevitable\\ blows, I think of those stones.\end{tabular}                                                                                                                                                \\ \hline\hline
\begin{tabular}[c]{@{}l@{}}Travel\\ Writing\end{tabular}                    & \multirow{4}{*}{Cliché}                                                                      & \begin{tabular}[c]{@{}l@{}}As the weekend comes to a close, reflect on the\\ kaleidoscope of experiences you've had, and \\ how each neighborhood has contributed to the \\ vibrant tapestry that is Los Angeles.\end{tabular}                                    & \begin{tabular}[c]{@{}l@{}}As the weekend comes to a close, you'll \\ have a new appreciation for the city's \\ diversity.\end{tabular}                                                                                                                      \\ \cline{1-1} \cline{3-4} 
\begin{tabular}[c]{@{}l@{}}Food\\ Writing\end{tabular}                      &                                                                                              & \begin{tabular}[c]{@{}l@{}}a dish that feels both familiar and unexpected, \\ much like discovering a hidden alley in a \\ well-known city. It's\end{tabular}                                                                                                     & \begin{tabular}[c]{@{}l@{}}a familiar dish with a surprising \\ twist. It's\end{tabular}                                                                                                                                                                     \\ \cline{1-1} \cline{3-4} 
\multirow{2}{*}{\begin{tabular}[c]{@{}l@{}}Literary\\ Fiction\end{tabular}} &                                                                                              & , watching his chest rise and fall with labored breaths                                                                                                                                                                                                           & -                                                                                                                                                                                                                                                            \\ \cline{3-4} 
                                                                            &                                                                                              & \begin{tabular}[c]{@{}l@{}}I noticed a vase of wilting flowers on the nightstand, \\ forgotten in the weight of more pressing concerns\end{tabular}                                                                                                               & The flowers on the nightstand had wilted                                                                                                                                                                                                                     \\ \hline
\end{tabular}
\vspace{2ex}
\caption{\label{qualllmedit} LLM edited spans from different categories across different sub-genres of writing }
\end{table*}

\subsection{Qualitative Insights into LLM edits}

Given how LLM-edited paragraphs are often preferred over default LLM-generated ones and sometimes even Writer edited paragraphs, we inspect if there are specific categories of edits where LLMs perform better, more so if that caters to a certain genre. Table \ref{qualllmedit} shows examples of original spans edited by LLMs that vary across genre and category. In general, there isn't a domain or category where LLM edits are better than the rest except for \texttt{Purple Prose}. To edit \texttt{Purple Prose} models mostly need to learn how to simplify the text (i.e., generate a paraphrase in simple language). LLMs are effective at Sentence Simplification \cite{dou-etal-2024-automatic} so this doesn't come across as a surprising finding. Categories like \texttt{Poor Sentence Structure} and \texttt{Unnecessary Exposition} deliver mixed results. Sometimes they are effective. For instance Row 7 in Table \ref{qualllmedit} shows how the model improves the structure by breaking the run-on very long sentence. However, for the other example in Row 6, the human edit is \textit{Then he hands his card, always leaving a 22\% tip} while the model edit is a mere paraphrase. Similarly, under the \texttt{Unnecessary Exposition} category, the LLM is very good at editing sentences with the following structure [Main clause] [comma] [exposition] where they simply remove the exposition (as learned from Writer edits). However, the bigger issue in this category is revealing the subtext. When a given input span doesn't conform to the structure above or consists of multiple sentences (Row 9, Table \ref{qualllmedit}) LLMs don't align with human edits. For \texttt{Cliché}, models mostly rewrite it in simpler language typically with fewer words which while being effective isn't what humans do. This also shows the fundamental overlap between \texttt{Cliché} and \texttt{Purple Prose} as shown in Section \ref{sec:writer_differences}. Finally, models are mostly ineffective at addressing \texttt{Lack of Specificity and Detail} (more details in section \ref{sec:mimic}). These findings demonstrate the challenge of editing text across any domains that require deeper emotional resonance or cultural commentary and the importance of rethinking the design of alignment for tasks with subjective or fuzzy rewards.

\vspace{-2ex}
\section{DISCUSSION}
\subsection{How is editing human writing different from LLM-generated text?}

Editing human writing and LLM-generated text presents distinct challenges and requires different approaches. Human writing often contains nuanced expressions, personal style, and contextual references that reflect the author's unique voice and experiences. Editors must preserve these elements while refining clarity, structure, and coherence. In contrast, LLM-generated text may lack consistent tone and exhibit repetitive patterns (Section \ref{data}). We asked writers to explain the differences in editing LLM-generated text compared to Human-written text. Several writers mentioned that LLM writing often required more extensive editing, mainly to remove unusual and sometimes nonsensical metaphors, inappropriate use of complex vocabulary that doesn't fit the context, and improving an overall tone that comes across as impersonal and mechanical. In an exchange, W3 noted ``\textit{I edit a lot of prose for my magazine but one thing that stuck with me as I was editing these paragraphs are the massive amount of cliché, histrionic descriptions, and direct exposition of intended meanings rather than effective representation. Indeed very strange}". They observed that the types of edits needed for LLM-generated text were often similar, but the sheer volume of necessary changes was higher than that of human writing. LLM-generated content's repetitive nature paradoxically made the human editors feel robotic while trying to improve it. By the edits that catered to deep emotional resonance or cultural commentary,

\subsection{How well can LLMs mimic edits from writers?} \label{sec:mimic}
Our preference ranking results in Section \ref{auto:rewrite} indicate that automatically edited paragraphs frequently rank second and sometimes first. This raises questions about LLMs' ability to analyze textual patterns and generate content closely resembling a given writer's edit. For the span {\color{red}``Janet lay in bed each night, her mind a whirlpool of restless thoughts"}, both LLM and writer identified it as cliché. The LLM edited it to {\color{ForestGreen}``Janet lay in bed each night, unable to sleep"} while the writer changed it to {\color{ForestGreen}``Each night, Janet lay prone in her bed and unable to sleep"}. LLMs can also split run-on sentences and improve poor structure. For {\color{red}``Sarah froze, realizing it was her high school sweetheart, Alex, whom she hadn't seen in over a decade"}, the LLM edited it to {\color{ForestGreen}``Sarah froze. It was Alex, her high school sweetheart. She hadn't seen him in over a decade"}, similar to the writer's edit. However, LLMs sometimes replace clichés with other clichés or fail to remove unnecessary exposition. The most challenging edit category is Lack of Specificity and Detail where LLMs often fail to add engaging details. For {\color{red}``Her irritation slowly morphed into a strange, disconnected calm"}, the writer added {\color{ForestGreen}``After all, the noise just meant that she wasn't the only one awake at this hour."} The model's edit was less effective: {\color{red}``Her irritation slowly morphed into a strange, disconnected calm. The repetitive thump-thump-thump became almost hypnotic, lulling her into a trance-like state"}. Much evocative detail in human writing comes from lived sensory experiences. LLMs on the other hand lack the grounded understanding that helps humans select vivid, emotionally resonant details. Additionally, when we combine autoregressive objectives with post-hoc adjustment through RLHF  ( typically designed to prevent toxic/harmful text generations) models often default to common, generic descriptions rather than specific ones since those are "safer" predictions. However, we note one potential limitation in our experiments is our reliance on few-shot instructions, requiring the model to learn rewriting from only a few examples. Training on the entire \corpus{} or more data might improve edit quality.

\subsection{What recommendations can we provide for future LLM-based writing support tools that aspire to improve the co-writing experience?}
Eminent author Curtis Sittenfeld calls LLM writing the literary equivalent of fat-free cookies \footnote{https://www.nytimes.com/2024/08/20/opinion/beach-read-ai.html}. LLMs are proficient at producing sentences that are grammatically correct and devoid of spelling errors. Beyond that, LLMs require extensive learning to effectively assist humans in improving their writing. In his essay \textit{Politics and English Language} \cite{orwell2013politics}, George Orwell said ``Never use a long word when a short one will do.'' LLM writing transgresses this simple rule by overusing lofty words. Clichés are bound to slip into even the best human writing, but when it comes to LLMs it simply cannot write without them. We believe this is partially a drawback of the technology behind LLMs. When an LLM calculates the probability of one word following another, clichés become very likely, because they've appeared so many times before. This explains why every other generated response is rife with clichés despite our prompt explicitly asking LLMs to avoid clichés and overused tropes (Table \ref{prompts}). LLMs need to learn how to identify and write without clichés such that it is engaging to every single reader. Overwriting is a bigger problem than underwriting. The rule for most writers is, ``If in doubt, cut it." \cite{nytedit} The Pulitzer Prize-winning writer John McPhee has called the process ``writing by omission." \cite{mcphee2015omission}. To become a better writer LLMs need to learn how to avoid unnecessary exposition. Last but not least, structure is what good writing hangs on \cite{nytedit}. Long, run-on sentences are hard to read, and LLMs need to know when and how to split effectively to better manage flow and clarity.

\subsection{What are the potential long-term effects on language evolution and writing styles as LLM becomes more prevalent and how can aligned editing tools help?}

The increasing prevalence of large language models (LLMs) could significantly impact language evolution and writing styles over time. There's potential for more homogenized writing as people rely on LLM-generated content, possibly leading to a reduction in linguistic diversity and individual voice \cite{shumailov2024ai}. However, well-designed editing tools aligned with expert writing practices could help counteract these effects. Such tools could encourage more nuanced and sophisticated language use, preserve stylistic diversity, and promote critical thinking about word choice and sentence structure. By highlighting elements of expert writing, these tools could elevate overall writing quality while still allowing for personal expression, potentially steering language evolution towards greater clarity, precision, and effectiveness in communication \cite{10.1145/3613904.3642625}.

\subsection{Why are there no significant differences in perceived writing quality or types of edits needed across texts generated by different large language models?}

\begin{figure}[!ht]
    \centering
        \includegraphics[width=0.5\textwidth]{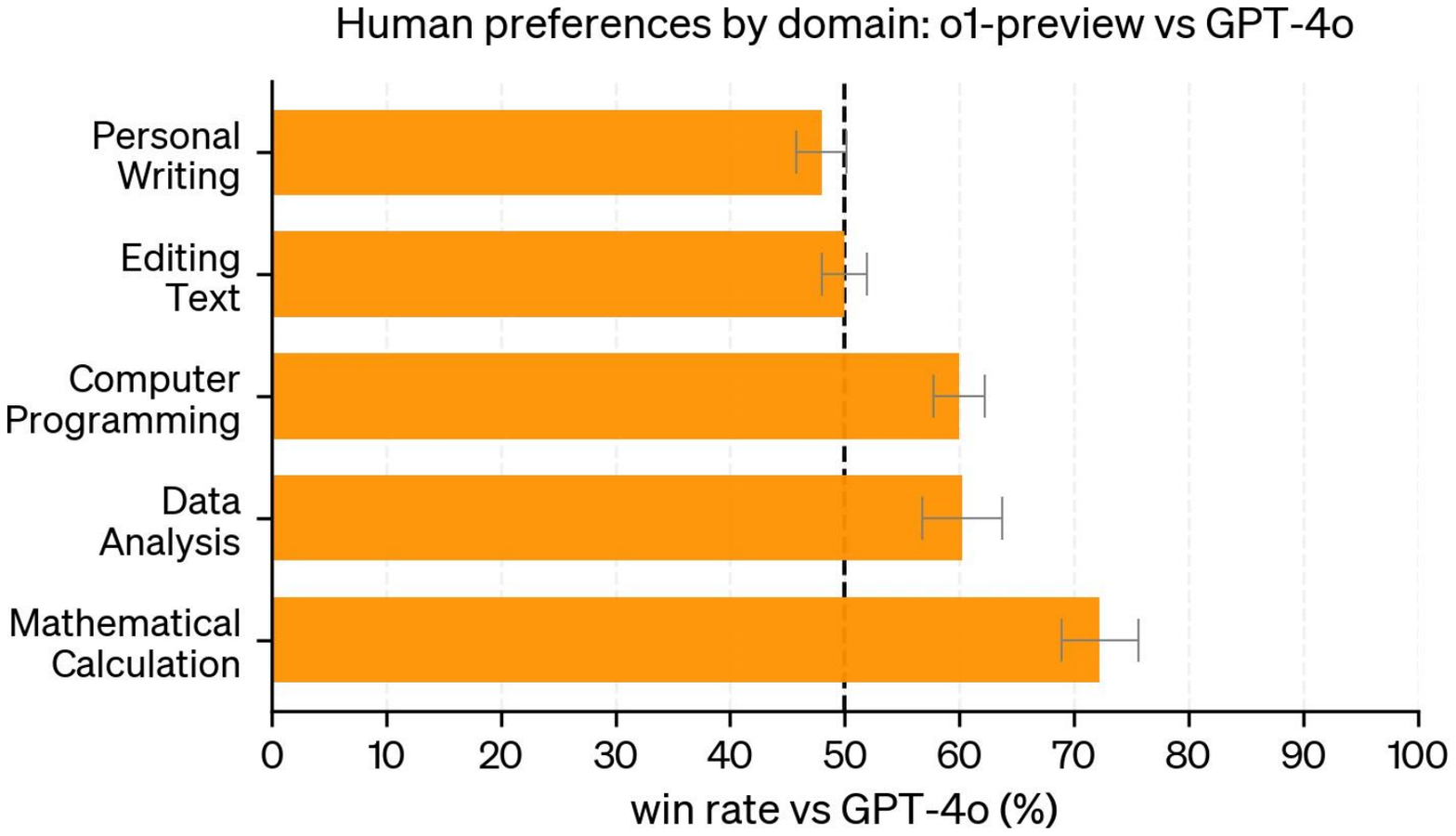}
        \vspace{-5ex}
    \caption{\label{ovso1} Preference Evaluation Results between GPT4o and GPT4-o1}
\end{figure}

The lack of significant quality differences between LLM writing is somewhat unexpected and warrants deeper examination. Several factors however may explain this surprising finding. Books play a crucial role in the training of generative AI systems. Their long, thematically consistent paragraphs provide information about how to generate coherent and fluent text. All three models (GPT-4, Claude 3.5, Llama 3.1) are pre-trained on Books3corpus \cite{kelly_books3_ai_2023} which constitute a bigger portion of Pile \cite{gao2020pile} pre-training data. Additionally, all LLMs rely on Scale AI for collecting preference data and there is very likely a significant overlap in workers who annotate the preferences as recently discussed \cite{wang2024scaleai}. Creative writing may also present an inherent "quality ceiling" for current LLM architectures, where all models encounter similar limitations in maintaining a consistent narrative voice and avoiding clichés—a pattern also observed by \cite{tian2024large} in their analysis of narrative generation. Our syntactic pattern analysis (Section 5.3) reveals remarkably similar templates and phrasings across models, further supporting the hypothesis of shared training foundations. While our methodology focused on concrete, actionable edit categories, we acknowledge that more subtle stylistic differences or subjective "vibes" may exist between models not captured by our current taxonomy \cite{dunlap2024vibecheck}. The consistency in writing quality across models ultimately points to broader questions about the current limitations of LLMs in creative writing tasks, rather than differences between specific implementations. Recent preference evaluation results from OpenAI's GPT4-o1 technical report \cite{openai2024learning} (See Figure \ref{ovso1}) corroborate our findings where there are no significant differences in GPT4o vs GPT4o1 for personal writing or editing text.

\section{Limitations}
While our study provides valuable insights into improving LLM-generated text through expert editing, there are several limitations to consider. Our study was conducted with 18 MFA-trained creative writers. While this ensured a high level of expertise, it may limit the generalizability of our findings. Future research could expand the participant pool to encompass diverse cultural backgrounds and writing traditions. The editing data primarily comes from literary fiction and creative non-fiction, making the identified idiosyncrasies and editing strategies potentially less applicable to other genres like technical writing, journalism, or scientific writing. Future work can expand on the line of work by including a broader range of writing styles and purposes. The selected LLMs (GPT-4, Claude 3.5 Sonnet, and Llama 3.1) are among the most advanced models, but they might not fully represent the entire spectrum of AI writing abilities. It should be noted that the evaluation of writing quality is inherently subjective, even with multiple annotators and inter-annotator agreement calculations. Experts may disagree on what constitutes an improvement, potentially influencing our results and their interpretation.Our automated methods for detecting and rewriting problematic spans relied on few-shot learning with a limited number of examples. While this approach showed promise, it may not fully capture the complexity and nuance of expert editing and training a model on the entire \corpus{} or additional data is required. 

It should be noted that while paragraph-level editing provides a balance between granularity and context, it may miss broader structural or thematic issues that become apparent only when considering longer pieces of writing. Last but not least, our study relied on professional writers editing AI-generated text for monetary compensation, which may have influenced the quality and nature of the edits. Editing one's own work typically involves more personal investment than editing text for pay, potentially leading to less motivation for substantial improvements \citep{byron2012rewards}. Additionally, the repetitive nature of editing multiple AI-generated paragraphs could lead to fatigue, especially if the content is perceived as uninteresting or lacking in creativity. This fatigue could result in less thorough or thoughtful edits as the task progresses. Finally, another potential limitation of LLM-based editing is the risk of hallucinated and factually inconsistent information being introduced during the editing process. We did not study hallucinations as part of our work, due to the chosen domain focus of fictional writing which has less stringent factuality requirements, yet prior work has documented that LLM-based text editing can introduce factual errors \cite{laban2023beyond}.

\section{Conclusion}
In this work, we present a comprehensive approach to mitigating idiosyncrasies and improving human-AI alignment in the writing process through expert editing. We i) develop a taxonomy of edit categories grounded in established writing practices, ii) create the LAMP corpus containing over 8,000 fine-grained edits by professional writers on LLM-generated text, and iii) design methods for automatic detection and rewriting of problematic spans. Our analysis reveals several key findings. Professional writers identify consistent categories of edits needed to improve AI writing. Surprisingly, there are no significant differences in perceived writing quality or types of edits needed across texts generated by different large language models (GPT-4, Claude 3.5, Llama 3.1). Automated methods using few-shot prompting can detect and rewrite problematic spans in LLM-generated text, though far from matching human expert performance. Finally, in terms of preference evaluations, writers consistently rank text edited by other writers highest, followed by LLM-edited text, with unedited LLM-generated text ranking lowest. As AI text generation becomes more prevalent, developing robust editing and alignment techniques will be crucial to ensure AI systems produce high-quality writing that meets human standards and enhances creativity and linguistic diversity.

\bibliographystyle{ACM-Reference-Format}
\bibliography{bibliography}

\appendix 
\section{Appendix}\label{sec:appendix}

\begin{figure*}
\centering
\includegraphics[width=0.9\textwidth]{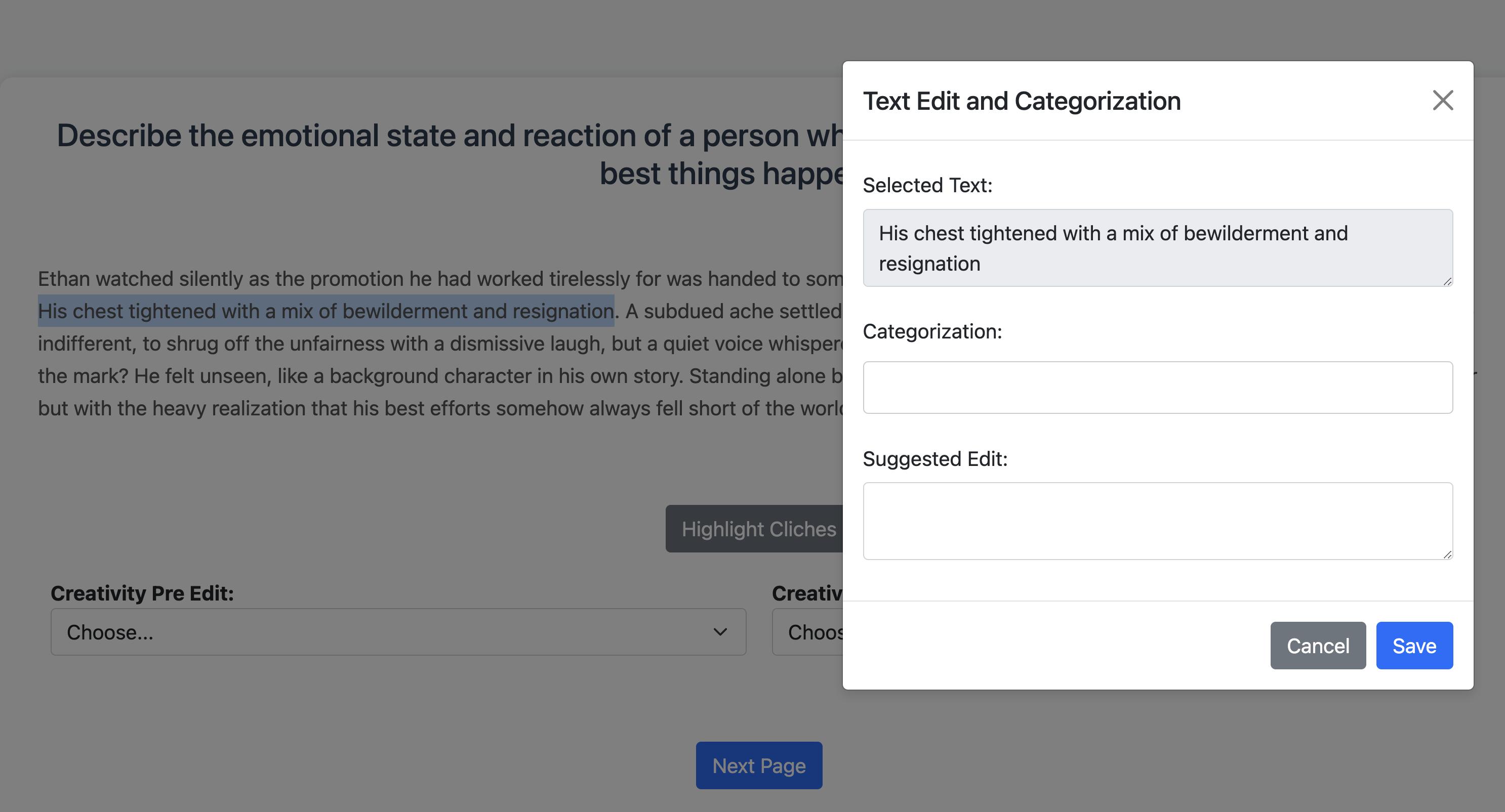}
\caption{\label{pilot}Interface for formative study to collect fine-grained labels}
\end{figure*}

\begin{table*}[!ht]
\small
\centering
\renewcommand{\arraystretch}{1.15}
\begin{tabular}{|l|l|}
\hline
\multirow{2}{*}{\begin{tabular}[c]{@{}l@{}}Instruction\\ Prompt\end{tabular}} & Summarize this paragraph into a single sentence open-ended question.\textbackslash{}n\textbf{ \{\{paragraph\}\}} \\ \cline{2-2} 
& Summarize this paragraph into a single sentence open-ended instruction.\textbackslash{}n\textbf{ \{\{paragraph\}\}}                                                                                                                                                                                                                                                                                                                                                                                                        \\ \hline
\begin{tabular}[c]{@{}l@{}}Response\\ Prompt\end{tabular}                     & \begin{tabular}[c]{@{}l@{}}Imagine you are a {\color{blue}fiction writer for the NewYorker}. Now write a paragraph (10-15 sentence) \\ as a response to the following question. Try your best to be original, avoiding clichés or\\ overused tropes. Do not use ornamental language and focus on nuance, simplicity, and subtext.\\ Start directly with your response. \textbackslash{}n\textbf{ \{\{instruction\}\}}\end{tabular}                                                                                                       \\ \hline
                                                                              & \begin{tabular}[c]{@{}l@{}}Imagine you are a {\color{blue}writer for the New York Times Modern Love section}. Now write a \\ paragraph (10-15 sentence) as a response to the following question. Try your best to be \\ original, avoiding clichés or overused tropes. Do not use ornamental language and focus\\ on nuance, simplicity, and subtext.Start directly with your response 
                                                                              \textbackslash{}n\textbf{ \{\{instruction\}\}}
                                    \end{tabular}                                                                                       \\ \hline
                                                                              & \begin{tabular}[c]{@{}l@{}}Imagine you are a {\color{blue}writer for the New York Times Cooking section}. Now write a paragraph\\ (10-15 sentence) as a response to the following question. Try your best to be original, avoiding\\ clichés or overused tropes. Do not use ornamental language and focus on nuance, simplicity, \\ and subtext.Start directly with your response 
                                                                              \textbackslash{}n\textbf{ \{\{instruction\}\}}
                                                                              
                                                                              \end{tabular}                                                                                            \\ \hline
                                                                              & \begin{tabular}[c]{@{}l@{}}Imagine you are a {\color{blue}writer for the New York Times Travel section}. Now write a paragraph\\ (10-15 sentence) as a response to the following question. Try your best to be original, avoiding\\ clichés or overused tropes. Do not use ornamental language and focus on nuance, simplicity, \\ and subtext.Start directly with your response 
                                                                              \textbackslash{}n\textbf{ \{\{instruction\}\}}
                                                                              \end{tabular}                                                                                             \\ \hline
                                                                              & \begin{tabular}[c]{@{}l@{}}Imagine you are a {\color{blue}beloved female Internet advice columnist whose trademark is deeply felt} \\{\color{blue} and frank responses grounded in your own personal experience}. Now write a paragraph \\ (10-15 sentence) as a response to the following question.Try your best to be original, avoiding \\ clichés or overused tropes. Do not use ornamental language and focus on nuance, \\ simplicity, and subtext.Start directly with your response 
                                                                              \textbackslash{}n\textbf{ \{\{instruction\}\}}
                                                                              
                                                                              \end{tabular} \\ \hline
\end{tabular}
\vspace{2ex}
\caption{\label{prompts}Prompts for generating instructions and responses}
\end{table*}

\newpage

\subsection{Idiosyncracy Span Detection Prompt} \label{app:detection_prompt}

\begin{figure*}[t]
\begin{myjsonblock}{Idiosyncracy Span Detection Prompt}
% \begin{multicols}{2}
% \onecolumn 
\small
You are given a paragraph of writing, and your goal is to provide feedback by selecting spans of text in the writing that could be improved and assign each problematic span to an error category. Below, we list the 7 error categories that you can choose from.\\
You are also provided 2 examples of paragraphs that were annotated by professional writers, which you can use to better understand the task and the error categories.\\
Error Categories:\\
- "Awkward Word Choice and Phrasing": Suggestions for better word choices or more precise phrasing to enhance clarity and readability.\\
- "Cliche": The use of hackneyed phrases or overly common imagery that lacks originality or depth.\\
- "Poor Sentence Structure": Feedback on the construction of sentences, recommending changes for better flow, clarity, or impact.\\
- "Unnecessary/Redundant Exposition": Redundant or non-essential parts of the text that could be removed/rephrased for conciseness.\\
- "Lack of Specificity and Detail": Need for more concrete details or specific information to enrich the text and make it more engaging.\\
- "Purple Prose": Identifying parts of the text that are seen as unnecessary ornamental and overly verbose.\\
- "Tense Consistency": Comments pointing out \\inconsistencies in verb tense that need to be\\ addressed for uniformity.\\
Example 1:
Input Text\\

Output:\\
Example Output in JSON format.\\
\\
Example 2:\\
(Similar to example 1)\\
Rules:\\
- Number of Spans -- You can provide feedback on \\multiple spans, and multiple spans can have the \\same category.\\
- Span must be verbatim -- The span you select must be \\verbatim from the paragraph, otherwise, the feedback \\will not be provided to the user.\\
- No Overlap -- Spans should not overlap, and one \\ span should not include the other.\\
- Single Category -- Each span should have exactly one\\ category from the categories listed above.\\
Paragraph:\\\\
PARAGRAPH
% \end{multicols}
\end{myjsonblock}
\end{figure*}

\clearpage

\subsection{Precision Metrics Explanation and Example} \label{sec:precision_explanation}

We illustrate the General and Categorical Precision on a simple example and justify the choice of the metric.

Imagine we have the following sentence, that has been annotated by a human annotator:
On this dark and stormy night, her heart skipped a beat as she was afraid of what was to come.
\begin{quote}
    ANNOTATION = 
    On this \textbf{dark and stormy night}\textsc{[CLICHÉ]}, her heart skipped a beat \textbf{as she was afraid of what was to come.}\textsc{[UNNECESSARY EXPOSITION]}
\end{quote}

Now let's imagine that System 1 and System 2 have produced the following predictions:

\begin{quote}
    SYSTEM 1 = 
    On this \textbf{dark and stormy night, her heart skipped a beat as she was afraid of what was to come.}\textsc{[CLICHÉ]}
\end{quote}

\begin{quote}
    SYSTEM 2 = 
    On this dark and \textbf{stormy night}\textsc{[CLICHÉ]}, her heart skipped a beat \textbf{as she was afraid of what was to come.}\textsc{[CLICHÉ]}
\end{quote}

We extract the annotated spans:
\begin{itemize}
    \item Span 1: characters [9,30]; category: CLICHÉ
    \item Span 2: characters [57, 94]; category: UNNECESSARY EXPOSITION
\end{itemize}

System 1 produced a single span:
\begin{itemize}
    \item Span 1: characters [9, 94]; category: CLICHÉ
\end{itemize}

System 2 produced two spans:
\begin{itemize}
    \item Span 1: characters [19,30]; category: CLICHÉ
    \item Span 2: characters [57, 94]; category: CLICHÉ
\end{itemize}

We can first compute General Precision, which disregards the category of the spans. It is the overlap between predicted spans and annotated spans, divided by the total amount of predicted characters:
\begin{itemize}
    \item General Precision (System 1) = ((30-9) + (94-57))  / (94-9) = 0.68
    \item General Precision (System 2) = ((30-19) + (94-57))  / ((30-19) + (94-57)) = 1.0
\end{itemize}

System 2 achieves a higher precision, as all the spans it predicted were included in the manual annotation. On the other hand, System 1 predicted a larger span that included the annotated span, but also additional characters, causing a lower precision score.

When consider Categorical Precision, overlap is only considered as valid if the overlapping spans coincide in category. The scores would be:
\begin{itemize}
    \item Categorical Precision (System 1) = (1*(30-9) + 0*(94-57))  / (94-9) = 0.25
    \item Categorical Precision (System 2) = (1*(30-19) + 0*(94-57))  / ((30-19) + (94-57)) = 0.23
\end{itemize}

System 1 achieved higher categorical precision by fully overlapping with the annotated CLICHÉ span, while System 2 only partially overlapped. Both systems incorrectly categorized the second span, resulting in lower precision scores. Precision scores can be inflated by reducing predictions, but our LLMs weren't instructed to optimize for precision. In fact, they tend to select more spans than human annotators, leading to high recall but potentially lower precision. We focus on precision to penalize systems that produce too many or overly large spans.

\subsection{Rewriting Prompts \label{rewriting_prompts}}
\begin{table*}[!ht]
\centering
\small
\begin{tabular}{|l|}
\hline
\begin{tabular}[c]{@{}l@{}}A cliché is a saying, idea, or element of an artistic work that has become overused to the point of losing its\\ original meaning or effect, even to the point of being weird, irritating, or bland\\ \\ You will be given example of 25 paragraphs with spans that count as Cliche and suggested edits that either\\  **REWRITES THE CLICHE or SIMPLY REMOVES IT**.\\ \\ Your task will then be to suggest edits (either spans or empty string) that gets rid of the cliche while making\\ the resulting paragraph coherent, given a new paragraph and highlighted span of Cliche from it. Do not simply\\ paraphrase or use fancy ornamental language; Try to keep each sentence short. Look at the examples carefully\\ \\ **IT IS VERY IMPORTANT TO MAKE SURE THAT YOUR EDITED TEXT ONCE ADDED TO THE\\ PARAGRAPH READS COHERENTLY AND GRAMMATICALLY CORRECT. \\ \\ For instance if you replace text within \textless{}span\textgreater{}\textless{}/span\textgreater tags with a longer span; please make sure the following\\ text after the edit, is its continuation. A simple way to ensure this is to ensure that the edited span has the same\\ casing and punctuation at the beginning and end as that of the original span.\\ \\ PLEASE FOLLOW THE OUTPUT SCHEMA AS THE EXAMPLES BELOW AND DO NOT RETURN\\ ANYTHING OTHER THAN THE EDITED SPAN WITHIN QUOTES\\ \\ Example 1\\ \\ Paragraph: Matthews had lived in the Valley all his life, and its rhythms and secrets were etched into his\\ being \textless{}span\textgreater{}like the lines on a well-worn map\textless{}/span\textgreater{}. He knew {[}...{]}\\ Original Span: "like the lines on a well-worn map"\\ Edited Span: "like creases in an old pocket map"\\ \\ .\\ .\\ .\\ \\ Example 18\\ \\ Paragraph: Husna sat at the ancient wooden {[}....{]} \textless{}span\textgreater{}The room was a bubble of quiet concentration, the\\ only sounds the clacking of the typewriter, the rustling of paper, and the occasional whistle of the teakettle in\\ the adjoining kitchen.\textless{}/span\textgreater\\ Original Span: "The room was a bubble of quiet concentration, the only sounds the clacking of the typewriter,\\ the rustling of paper, and the occasional whistle of the teakettle in the adjoining kitchen."\\ Edited Span: "The room was quiet. The outside world did not exist. At times, Husna tapped her foot. Shah \\ Sahib coughed and she would stop. The typewriter never did."\\ .\\ .\\ \\ \\ Example 25\\ \\ Paragraph: Last night, I dreamt of an {[}....{]} She didn't speak, but her eyes \textless{}span\textgreater{}communicated a haunting mix\\ of sadness and knowing, as if she held\textless{}/span\textgreater the weight of forgotten secrets. I felt a {[}...{]}\\ Original Span: "communicated a haunting mix of sadness and knowing, as if she held"\\ Edited Span: "conveyed"\end{tabular} \\ \hline
\end{tabular}
\vspace{2ex}
\caption{Prompt to rewrite Cliche}
\vspace{-2ex}
\end{table*}

\begin{table*}[!ht]
\centering
\small
\begin{tabular}{|l|}
\hline
\begin{tabular}[c]{@{}l@{}}Poor sentence structure refers to writing that is difficult to understand or lacks clarity due to issues with how\\ sentences are constructed. It encompasses issues like run-on sentences, fragments, misplaced or dangling \\ modifiers, lack of variety, overuse of passive voice, improper parallelism, and unclear pronoun references, \\ all of which impede clear communication and reader comprehension\\ \\ You will be given examples of 25 paragraphs with text within \textless{}span\textgreater{}\textless{}/span\textgreater tags that shows poor sentence\\  structure and suggested edits that either **REWRITES WITH IMPROVED SENTENCE STRUCTURE**.\\ \\ Your task will then be to suggest edits that rewrite the text within the span tags with better sentence structure \\ while making the resulting paragraph coherent, given a new paragraph and highlighted span of poor sentence \\ structure from it. Do not use fancy ornamental language; Look at the examples carefully and do not output\\  anything after closing quotes.\\ \\ **IT IS VERY IMPORTANT TO MAKE SURE THAT YOUR EDITED TEXT ONCE ADDED TO THE \\ PARAGRAPH READS COHERENTLY AND GRAMMATICALLY CORRECT. For instance, if you replace \\ text within \textless{}span\textgreater{}\textless{}/span\textgreater tags with a longer span; please make sure the following text after the edit, is its \\ continuation.\\ \\ PLEASE FOLLOW THE OUTPUT SCHEMA AS THE EXAMPLES BELOW AND DO NOT RETURN\\ ANYTHING OTHER THAN THE EDITED SPAN WITHIN QUOTES\\ .\\ .\\ .\\ Example 4\\ \\ Paragraph: \textless{}span\textgreater{}As the night wore on, Z.'s laughter grew louder, his words slurring together like a sloppy\\ melody. N. and I exchanged a knowing glance, our concern simmering beneath the surface.\textless{}/span\textgreater At first, it\\ was just a slight stumble, a misstep that could be brushed off as a joke. {[}.....{]}\\ \\ Original Span: "As the night wore on, Z.'s laughter grew louder, his words slurring together like a sloppy melody.\\ N. and I exchanged a knowing glance, our concern simmering beneath the surface."\\ Edited Span: "Z. was drinking more and more as the night went on. He laughed more loudly. His words started to\\ slur, blurring one into the next. I looked at N., who knew what I was thinking. We were going to have to take \\ care of him.".\\ .\\ .\\ \\ Example 13\\ \\ Paragraph: \textless{}span\textgreater{}As I step into the quiet, garden-facing room on the second floor, I'm struck by the sense of stillness \\ that pervades the space\textless{}/span\textgreater{}. The occupants, an elderly couple, sit motionless in their armchairs, their {[}....{]}\\ Original Span: "As I step into the quiet, garden-facing room on the second floor, I'm struck by the sense of \\ stillness that pervades the space"\\ Edited Span: "A sense of stillness pervades the garden-facing room on the second floor" .\\ .\\ .\\ \\ Example 25\\ \\ Paragraph: Chef Amelia  raced {[}.....{]} \textless{}span\textgreater{}She plastered on a polite smile, determined not to let her personal history\\ interfere with her professional duties.\textless{}/span\textgreater As Daniel approached, plate in hand, Amelia steeled herself {[}.....{]}\\ Original Span: "She plastered on a polite smile, determined not to let her personal history interfere with her \\ professional duties."\\ Edited Span: "She shot a dutiful smile for anyone who was looking. This was an important night, and she wasn't\\  going to let the past get in the way of a job well done."\end{tabular} \\ \hline
\end{tabular}
\vspace{2ex}
\caption{Prompt to rewrite Poor Sentence Structure}
\vspace{-2ex}
\end{table*}

\begin{table*}[!ht]
\centering
\small
\begin{tabular}{|l|}
\hline
\begin{tabular}[c]{@{}l@{}}Unnecessary or redundant exposition in writing refers to providing excessive explanatory information that \\ doesn't contribute meaningfully to the story, characters, or overall narrative.\\ \\ You will be given example of 25 paragraphs with text within \textless{}span\textgreater{}\textless{}/span\textgreater tags that count as \\ unnecessary/redundant exposition and suggested edits that either **REWRITES IT IN FEWER WORDS\\ or SIMPLY REMOVES IT**.\\\\ Your task will then be to suggest edits that rewrites the text within the span tags correcting the unnecessary\\ /redundant exposition while making the resulting paragraph coherent, given a new paragraph and \\ highlighted text within of unnecessary/redundant exposition. Do not simply paraphrase or use fancy \\ ornamental language or repeat the same thing in the edited span; Look at the examples carefully.\\ \\ **IT IS VERY IMPORTANT TO MAKE SURE THAT YOUR EDITED TEXT ONCE ADDED TO THE \\ PARAGRAPH READS COHERENTLY AND GRAMMATICALLY CORRECT. \\ \\ For instance if you replace text within \textless{}span\textgreater{}\textless{}/span\textgreater tags with a shorter span; please make sure the\\ following text after the edit, is its continuation. Simple way to ensure this is to make sure that the edited\\ span has the same casing and/or punctuation at the beginning and end as that of the original span.\\ \\ PLEASE FOLLOW THE OUTPUT SCHEMA AS THE EXAMPLES BELOW AND DO NOT RETURN\\ ANYTHING OTHER THAN THE EDITED SPAN WITHIN QUOTES \\ \\ .\\ ,\\ Example 2\\ Paragraph: In spring, when the first buds unfurled {[}...{]} embrace of varenyky dinners provided comfort against \\ the chill \textless{}span\textgreater{}, each bite narrating a history of resilience and hope\textless{}/span\textgreater{}. It was through {[}...{]}\\ Original Span: ", each bite narrating a history of resilience and hope"\\ Edited Span: ""\\ .\\ .\\ .\\ \\ Example 18\\ \\ Paragraph: \textless{}span\textgreater{}As Oghi watched his mother-in-law, Mrs. Kim, he felt a subtle sense of unease settle in the\\ pit of his stomach.\textless{}/span\textgreater It wasn't just the uncharacteristic behavior itself - {[}...{]}\\ Original Span: "As Oghi watched his mother-in-law, Mrs. Kim, he felt a subtle sense of unease settle in the pit \\ of his stomach."\\ Edited Span: "Oghi watched his mother-in-law Mrs. Kim with heightening unease."\\ \\ .\\ .\\ \\ \\ Example 23\\ \\ Paragraph: The small room {[}....{]} They teased and corrected each other's recollections \textless{}span\textgreater , creating a tapestry \\ of resilience and camaraderie\textless{}/span\textgreater{}.It wasn't all smooth-sharp words resurfaced around old wound, {[}....{]}\\ Original Span: ", creating a tapestry of resilience and camaraderie"\\ Edited Span: ""\\ \\ .\\ .\end{tabular} \\ \hline
\end{tabular}
\vspace{2ex}
\caption{Prompt to rewrite Unnecessary or redundant exposition}
\end{table*}

\begin{table*}[!ht]
\centering
\small
\begin{tabular}{|l|}
\hline
\begin{tabular}[c]{@{}l@{}}Lack of Specificity and Detail in writing refers to the absence of concrete and specific information, which can make the text\\ feel vague and unengaging. The need for more concrete details or specific information is crucial to enrich the text and make\\ it more engaging. Specificity helps to create vivid imagery, provides clarity, and connects with the reader on a deeper level.\\ doesn't contribute meaningfully to the story, characters, or overall narrative.\\ \\ You will be given example of 25 paragraphs with text within \textless{}span\textgreater{}\textless{}/span\textgreater tags that lacks specificity and detail and suggested\\  edits that either **REWRITES WITH SPECIFICITY AND DETAIL**.\\ \\ \\ Your task will then be to suggest edits that rewrites the text within the span tags with specificity and detail that is engaging while\\ making the resulting paragraph coherent, given a new paragraph and highlighted span of lack of specificity and detail from it. \\ Do not simply paraphrase or use fancy ornamental language; Look at the examples carefully and do not output anything after\\ closing quotes.\\ \\ \\ **IT IS VERY IMPORTANT TO MAKE SURE THAT YOUR EDITED TEXT ONCE ADDED TO THE PARAGRAPH READS \\ COHERENTLY AND GRAMMATICALLY CORRECT. For instance if you replace text within \textless{}span\textgreater{}\textless{}/span\textgreater tags with a longer\\ span; please make sure the following text after the edit, is its continuation. Simple way to ensure this is to make sure that the edited \\ span has the same casing and punctuation at the beginning and end as that of the original span.\\ \\ PLEASE FOLLOW THE OUTPUT SCHEMA AS THE EXAMPLES BELOW AND DO NOT RETURN\\ ANYTHING OTHER THAN THE EDITED SPAN WITHIN QUOTES \\ \\ Example 1\\ Paragraph: Sarah Mitchum's marriage appeared outwardly conventional, but subtle tensions simmered beneath the surface. She \\ and {[}.....{]} leaving Sarah feeling increasingly isolated within her \textless{}span\textgreater{}own marriage.\textless{}/span\textgreater{}\\Original Span: "within her own marriage."\\ Edited Span: ".Their marriage had run its course. There was no coming back."\\ .\\ .\\ .\\ Example 15\\ \\ Paragraph: \textless{}span\textgreater{}Dr. Arthur Steiger's fall from grace began with a series of whispered concerns among his colleagues at \\ Cormac General Hospital.\textless{}/span\textgreater{}The small-town pain specialist had always been known {[}....{]}\\ Original Span: "Dr. Arthur Steiger's fall from grace began with a series of whispered concerns among his colleagues at Cormac \\ General Hospital."\\ Edited Span: "Pain was Dr. Arthur Steiger's forte. Not inflicting it, that is, but resolving it. Whenever a patient had problem, \\ whether a tear in atendon, a sprain, a knock, a headache, a broken bone– it was Dr. Steiger that knew what to do."\\ \\ .\\ .\\ Example 21\\ \\ Paragraph: Mila sat on her porch a week after the storm had hit, sipping lukewarm tea. {[}....{]} Each night \textless{}span\textgreater{}it grew louder, \\shifting from a whisper to a groan, but she had dismissed it, too tired from long days at work\textless{}/span\textgreater{}. {[}.....{]}\\ Original Span: "it grew louder, shifting from a whisper to a groan, but she had dismissed it, too tired from long days at work"\\ Edited Span: "lying like blanched spinach in her IKEA bed, trying not to think about another day of writing emails with someone\\ else's  signature on them and pretending not to care what John Blanchett, CEO of Executive Industries thought of her blouse--in \\other words,another day as John's executive assistant--"\\ \\ \\ .\\ .\end{tabular} \\ \hline
\end{tabular}
\vspace{2ex}
\caption{Prompt to rewrite Lack of Specificity and Detail}
\end{table*}

\begin{table*}[!ht]
\centering
\small
\begin{tabular}{|l|}
\hline
\begin{tabular}[c]{@{}l@{}}In literary criticism, purple prose is overly ornate prose text that may disrupt a narrative flow by drawing undesirable attention\\ to its own extravagant style of writing, thereby diminishing the appreciation of the prose overall. Purple prose is characterized\\  by the excessive use of adjectives, adverbs, and metaphors.\\ \\ You will be given example of 25 paragraphs with text within \textless{}span\textgreater{}\textless{}/span\textgreater tags that has purple prose in it and suggested edits\\ that either **REWRITES THEM WITH SIMPLER WORDS OR REMOVES IT**.\\ \\ Your task will then be to suggest edits that rewrites the text within the span tags altering the purple prose while making the\\  resulting paragraph coherent, given a new paragraph and highlighted span of purple prose from it. Do not simply paraphrase\\  or use fancy ornamental language; Look at the examples carefully and do not output anything after closing quotes.\\ \\ **IT IS VERY IMPORTANT TO MAKE SURE THAT YOUR EDITED TEXT ONCE ADDED TO THE PARAGRAPH \\ READS COHERENTLY AND GRAMMATICALLY CORRECT. For instance if you replace text within \textless{}span\textgreater{}\textless{}/span\textgreater \\ tags with a longer span; please make sure the following text after the edit, is its continuation. Simple way to ensure this is \\ to make sure that the edited span has the same casing and punctuation at the beginning and end as that of the original span. \\ \\ PLEASE FOLLOW THE OUTPUT SCHEMA AS THE EXAMPLES BELOW AND DO NOT RETURN ANYTHING\\  OTHER THAN THE EDITED SPAN WITHIN QUOTES \\ \\ .\\ .\\ Example 2\\ \\ Paragraph: \textless{}span\textgreater{}Fruto never intended to stir anything beyond the melting pot of their weekly card game.\textless{}/span\textgreater But \\ when the chatter turned to the dry monotony of their jobs, Fruto found himself blurting out,  {[}....{]}\\Original Span: "Fruto never intended to stir anything beyond the melting pot of their weekly card game."\\ Edited Span: "Fruto hadn't meant to disrupt the routine of their weekly card game."\\ .\\ .\\ .\\ \\ Example 16\\ \\ Paragraph: My mother cried,  {[}....{]} All of it vanished\textless{}span\textgreater{}, cycling back through her mind, not as numbers but memories \\ of scraped knees she bandaged alone and birthdays where her absence was felt more acutely than her presence. The sobs \\ emerged from this deep well of unspoken expectations, leaving behind a residue of weary resilience and a few hopeful \\ echoes yet unwilling to completely extinguish.\textless{}/span\textgreater\\ Original Span: ", cycling back through her mind, not as numbers but memories of scraped knees she bandaged alone and \\ birthdays where her absence was felt more acutely than her presence. The sobs emerged from this deep well of unspoken\\  expectations, leaving behind a residue of weary resilience and a few hopeful echoes yet unwilling to completely extinguish."\\ Edited Span: "She cried. She cried deep from this well of scraped knees she bandaged alone and birthdays she missed to \\ work. She cried for unfairness. She cried without relief." .\\ .\\ \\ \\ Example 24\\ \\ Paragraph: \textless{}span\textgreater{}As they navigated their final year of high school, Maya and Jake found themselves at a crossroads, their\\ educational paths diverging like tributaries of a river.\textless{}/span\textgreater {[}....{]}\\ Original Span: "As they navigated their final year of high school, Maya and Jake found themselves at a crossroads, their \\ educational paths diverging like tributaries of a river."\\ Edited Span: "The final year of high school was pulling Maya and Jake in different directions."\\ .\\ .\end{tabular} \\ \hline
\end{tabular}
\vspace{2ex}
\caption{Prompt to rewrite Purple Prose}
\end{table*}

\end{document}